%% file: main.tex
\documentclass{article}

\usepackage{microtype}
\usepackage{graphicx}
\usepackage{booktabs} 

\usepackage{hyperref}

\usepackage[accepted]{icml2026}

\usepackage{amsmath}
\usepackage{amssymb}
\usepackage{mathtools}
\usepackage{amsthm}

\usepackage[capitalize,noabbrev]{cleveref}

\usepackage{multirow}
\usepackage{color}
\usepackage{fancybox}
\usepackage{mathrsfs}  
\usepackage{makecell}  
\usepackage{enumitem}  

\usepackage{algorithmic}

\newcommand{\mS}{\Sigma} 
\newcommand{\mC}{\mathcal{C}} 
\newcommand{\mD}{\mathcal{D}} 
\newcommand{\mM}{\mathcal{M}} 
\newcommand{\mMpt}{\mM_\mathrm{pretrain}}
\newcommand{\mV}{\mathcal{V}} 
\newcommand{\dmodel}{d_{\mathrm{model}}} 
\newcommand{\mStrain}{\mS_\mathrm{train}}
\newcommand{\mSpt}{\mS_\mathrm{pretrain}}

\newcommand{\Snew}{S_\mathrm{new}}
\newcommand{\Sho}{S_\mathrm{held\text{-}out}}

\DeclareMathOperator*{\argmin}{\arg\!\min}

\usepackage{pifont}    
\usepackage{xcolor}
\newcommand{\cmark}{\textcolor{green!60!black}{\ding{51}}}
\newcommand{\xmark}{\textcolor{red!60!black}{\ding{55}}}

\usepackage{listings}
\usepackage{tcolorbox}
\tcbuselibrary{skins,breakable}
\usepackage{fontawesome5}

\definecolor{custom_green}{HTML}{60B08C}

\tcbset{
 highlightbox/.style={
  enhanced,
  breakable,
  colback=custom_green!5,
  colframe=custom_green!50,
  boxrule=0pt,
  leftrule=2.5pt,
  sharp corners,
  left=8pt,
  right=6pt,
  top=3pt,
  bottom=3pt,
  coltitle=black,
  attach boxed title to top left={yshift=-2mm, xshift=3mm},
  boxed title style={
    colback=white,
    colframe=white,
    sharp corners,
    boxrule=0pt,
    left=3pt,
    right=3pt,
    top=0pt,
    bottom=0pt
  },
  }
}

\newtcolorbox{highlight}[2][]{highlightbox,title=#2,#1}

\theoremstyle{plain}
\newtheorem{theorem}{Theorem}[section]

\theoremstyle{definition}
\newtheorem{definition}[theorem]{Definition}

\theoremstyle{remark}

\usepackage[textsize=tiny]{todonotes}

\begin{document}

\twocolumn[
\icmltitle{Skill Neologisms: Towards Skill-based Continual Learning}
\icmlsetsymbol{equal}{*}

\begin{icmlauthorlist}
\icmlauthor{Antonin Berthon}{cam}
\icmlauthor{Nicolas Astorga}{cam}
\icmlauthor{Mihaela van der Schaar}{cam}
\end{icmlauthorlist}

\icmlaffiliation{cam}{University of Cambridge}

\icmlcorrespondingauthor{Antonin Berthon}{armb3@cam.ac.uk}

\icmlkeywords{Large Language Models, Skills Neologisms, Skill Composition, Skill-based Continual Learning} 

\vskip 0.3in
]

\printAffiliationsAndNotice{}  
\begin{abstract}
\input{sections/00_abstract}
\end{abstract}

\input{sections/01_introduction}

\input{sections/02_formalism}

\input{sections/03_methods}

\input{sections/04-motivating_exp}

\input{sections/05_experiments}

\input{sections/06_discussion}

\section*{Impact Statement}
This paper presents work whose goal is to advance the field of machine learning. Skill neologisms are a method for extending LLM capabilities without weight updates, and their societal implications are broadly similar to those of LLMs and parameter-efficient fine-tuning methods more generally. We do not identify specific consequences that we feel need to be highlighted here.

\section*{Acknowledgments}
This work was supported by Azure sponsorship credits granted by Microsoft’s AI for Good Research Lab. AB's research is supported by funding from Eedi, and NA is sponsored by
W.D. Armstrong Trust.

\bibliography{references}
\bibliographystyle{icml2026}

\newpage
\appendix
\onecolumn

\counterwithin{figure}{section}
\counterwithin{table}{section}
\renewcommand\thefigure{\thesection\arabic{figure}}
\renewcommand\thetable{\thesection\arabic{table}}

\input{sections/app_00_main}

\end{document}

%% file: sections/00_abstract.tex
Modern LLMs show mastery over an ever-growing range of skills, as well as the ability to compose them flexibly. However, extending model capabilities to new skills in a scalable manner is an open problem: fine-tuning and parameter-efficient variants risk catastrophic forgetting, while context-based approaches have limited expressiveness and are constrained by the model's effective context. 
We explore \textit{skill neologisms}--soft tokens integrated in the model's vocabulary and optimized to improve capabilities over a specific skill--as a way to selectively acquire new skills without weight updates.
We first observe that pretrained LLMs already exhibit tokens associated with procedural knowledge.
We then show on a controlled synthetic task that skill neologisms can be learned to improve model capabilities on specific skills while being composable with out-of-distribution skills, and that independently trained skill neologisms can be composed zero-shot. 
Finally, we validate zero-shot composition of independently learned skill neologisms on the more realistic natural language setting of the Skill-Mix benchmark~\citep{yu2024skill}.
These results suggest that skill neologisms may provide a scalable path towards skill-based continual learning. 

%% file: sections/01_introduction.tex
\section{Introduction}
Recent works have established that pretrained LLMs develop mastery over various skills and the ability to combine them beyond the pretraining distribution~\citep{arora2023theory, chen2023skills, yu2024skill}. As LLMs are used to tackle an ever-growing range of problems, the ability to continuously grow model capabilities to new skills in a scalable fashion is a promising research direction.

\begin{figure}[h]
\centering
\includegraphics[width=1\linewidth]{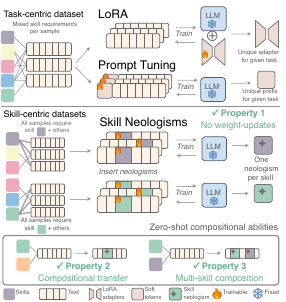}
\caption{\textbf{Overview of Skill Neologisms.}}
\label{fig:fig1}
\end{figure}

Yet, existing approaches to extend model capabilities fall short of this objective (Table~\ref{tab:comparison}).
Fine-tuning models on new datasets risks catastrophic forgetting~\citep{kirkpatrick2017overcoming, luo2025empirical}, where previously mastered capabilities might disappear and safety risks might be introduced~\citep{qi2023fine}. In-context learning has shown some success at skill composition in simple settings~\citep{chen2023skills, levy2023diverse, xu2024do}, but it does not adapt as well as parameter-efficient fine-tuning (PEFT) methods~\citep{liu2022few}, and does not scale because of effective context limitations~\citep{hsieh2024ruler}.
Prompt tuning~\citep{lester2021power} can adapt models to new tasks by only learning soft tokens prepended to the prompt, and has been shown to rival full fine-tuning in some settings~\citep{genewein2025understanding}. However, prefixes are typically task-specific instead of skill-specific, and learned prefixes cannot be composed or adapted to new settings without retraining~\citep{asai2022attempt, wang2023multitask}.

\begin{highlight}{\textbf{Central question}}
Can the compositional abilities of LLMs be leveraged to learn new composable skills without weight updates?
\end{highlight}

\begin{table}[h]
\centering
\caption{Comparison of different approaches for skill-based continual learning. \textsuperscript{‡}E.g., full fine-tuning and LoRA \citep{hu2022lora}. \textsuperscript{§}E.g., prompt tuning \citep{lester2021power} and related methods. *Achievable with skill-centered training.}
\label{tab:comparison}
\resizebox{\linewidth}{!}{\input{tables/comparison}}
\end{table}

We term this objective \textit{skill-based continual learning} and distinguish the following required properties: ($\triangleright$ \textbf{P1}) New skills can be learned without modifying model parameters; ($\triangleright$ \textbf{P2}) Learned skills are composable with other existing skills, including ones out-of-distribution from the training set; ($\triangleright$ \textbf{P3}) Multiple skills learned independently can be composed without joint training. 

In this work, we investigate whether \textbf{skill neologisms} might enable these properties (Figure~\ref{fig:fig1}). Inspired by neologisms proposed by~\citet{hewittposition} for human-machine communication, skill neologisms aim to learn new vocabulary elements that, when provided in the model's context, improve the model capabilities on a specific skill. While sharing prompt tuning's core mechanism of optimizing soft tokens on a frozen model (P1), skill neologisms differ in two ways that are critical for composability:
\begin{itemize}[leftmargin=3mm, itemsep=0pt, topsep=0pt]
\item \textbf{Skill-centered training}. Training uses datasets where every sample requires the target skill, mixed with diverse skills already mastered by the model. Such datasets can be constructed in many settings, for example by leveraging the metacognitive capabilities of modern LLMs~\citep{didolkar2024metacognitive} (see Section~\ref{sec:skill-centered_dataset}).
\item \textbf{Vocabulary-level integration}. Individual skills are learned via soft tokens (\textit{skill tokens}) integrated in the model vocabulary, allowing the model to interact with the skill's procedural knowledge via in-context learning.
\end{itemize}
These two components encourage learning generally composable skill representations (P2), and enable zero-shot composition of independently learned skills (P3).

Our main contributions are as follows:
\begin{itemize}[leftmargin=3mm, itemsep=0pt, topsep=0pt]
\item We propose skill neologisms as a path toward skill-based continual learning (\S~\ref{sec:method}), and motivate this approach via empirical evidence that pretrained LLMs naturally exhibit vocabulary elements that encapsulate procedural knowledge (\S~\ref{sec:xor}).
\item We demonstrate in controlled settings (\S~\ref{sec:result_main}) that skill neologisms compose with OOD skills unseen during training (P2) and enable zero-shot composition of independently learned skills (P3).
\item We provide ablation experiments (\S~\ref{sec:result_ablation}) analyzing how token capacity and composition complexity in the training set affect learning of composable skill representations.
\item We validate zero-shot composition of independently learned skill neologisms on the Skill-Mix benchmark~\citep{yu2024skill}, demonstrating their wider applicability to more realistic natural language settings (\S~\ref{sec:skill-mix}).
\end{itemize}

Code to reproduce our experiments can be found at {\small\url{https://github.com/antoninbrthn/skill-neologisms}}.

%% file: tables/comparison.tex
\begin{tabular}{lccc}
\toprule

 & \textbf{P1: No Weight} & \textbf{P2: Composes} & \textbf{P3: Multi-Skill} \\
 \textbf{Method} & \textbf{Updates} & \textbf{w/ OOD Skills } & \textbf{Composition} \\
\midrule
Fine-tuning-based\textsuperscript{‡} & \xmark & \xmark & \xmark \\
Prefix-based\textsuperscript{§} & \cmark & \cmark* & \xmark \\
\textbf{Skill neologisms} & \cmark & \cmark & \cmark \\
\bottomrule
\end{tabular}

%% file: sections/02_formalism.tex
\section{Preliminaries} 
\label{sec:formalism}
\subsection{Skills and Composition in Large Language Models} 
We build on the formalism introduced in~\citet{arora2023theory} in which skills refer to procedural knowledge—reusable capabilities such as arithmetic operations or logical reasoning—rather than factual knowledge. In this framework, any piece of text t is related to a set of skills S, and the understanding of text t requires mastery over all its underlying skills as well as their composition.

Given a set of skills $\mS$, we denote by $\mC_k(\mS)$ the set of text pieces that require a $k$-tuple of skills from $\mS$. By extension, $\mC_k(S_1,., S_i, \mS)$ denotes text pieces that require \textit{at least} skills $S_1$, ., $S_i$, mixed with $k-i$ other skills from $\mS$. 

\textbf{Closed-form assumption} What does it mean to understand a text snippet $t$? A key assumption from~\citet{arora2023theory} is that the understanding of any piece of text can be tested via closed-form questions. This might be trivial if $t$ relates to a closed-form question (eg \textit{"find the following number: 1,2,3,5,8,.."}), or by generating a set of multiple-choice questions as described in~\citet{arora2023theory}. 

\textbf{Composition beyond training}
Modern LLMs demonstrate the ability to understand combinations of skills beyond their training distribution~\citep{wei2022emergent, he2024learning, yu2024skill, zhao2024can}. Theoretical analysis presented in \citet{arora2023theory} links the emergence of skill composition ability to model scaling. Namely, scaling up model parameters by an order of magnitude leads to the same level of competence on $2k$-tuples of skills as the competence on $k$-tuples of the original model.

\subsection{Soft Prompts and Prompt Tuning} 
\textbf{Soft tokens} 
Soft tokens $s=(s_1, ..., s_l)$ are sequences of continuous vectors of size $\dmodel$ (matching the model's hidden dimension), that can be inserted in a model's context after skipping the embedding matrix.

\textbf{Prompt tuning} 
Prompt-tuning~\citep{lester2021power} is a parameter-efficient fine-tuning approach where trainable soft tokens $s$ are introduced as a prefix to the model's continuous representation of the input context. Only the soft tokens are optimized on a training set by back-propagating through the model while keeping the model's parameters frozen.

\textbf{Expressivity of prompt tuning}
Recent works~\citep{petrov2024prompting, genewein2025understanding} study the conditions under which methods like prompt tuning might or might not succeed at learning a new task. Informally, a necessary condition is that the new task is \textit{not too different} from tasks within the model's pretraining distribution, so that the model weights contain the necessary circuits to solve the new task.

\subsection{Vocabulary Extensions via Neologisms}
Neologisms embedding learning~\citep{hewittposition} uses learnable soft tokens as new vocabulary elements in a model's tokenizer and embedding matrix, that can then be used in prompts alongside text tokens. We denote a neologism of length $l$ as soft tokens $s=(s_1, ..., s_l)$, which extend the model's embedding matrix to $E'=E \cup {s} \in \mathbb{R}^{\dmodel \times (|\mathcal{V}|+l)}$ with columns $(s_1, ..., s_l)$, and its vocabulary to $l$ tokens: $\mV' = \mV \cup \{\langle S_1\rangle, ..., \langle S_l\rangle \}$.

Like prompt tuning, the soft tokens of a neologism can be trained on samples that include the neologism tokens by back-propagating gradients through the frozen model. This is done via preference-based learning in~\citet{hewittposition} but can be done similarly with supervised fine-tuning (SFT) or Reinforcement Learning Fine-tuning (RLFT).

%% file: sections/03_methods.tex
\section{Skill-based Continual Learning via Skill Neologisms}
\label{sec:method}

\begin{figure*}[t!]
\centering
\includegraphics[width=1\linewidth]{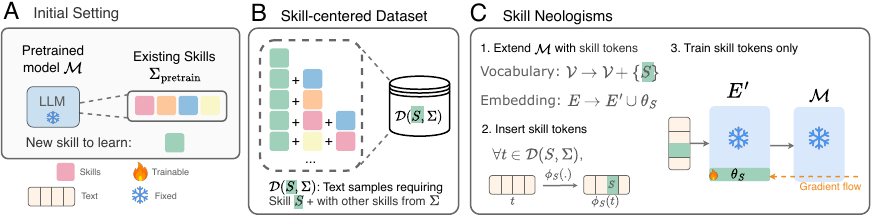}
\vspace{-20pt}
\caption{\textbf{Overview of Skill Neologisms.} (A) We consider pretrained model endowed with a set of implicit skills learned during pretraining. (B) A \textit{skill-centered dataset} contains snippets of text that require \textit{at least} the skill of interest, composed with pretraining skills. (C) \textit{Skill neologisms} append new token embeddings to the model's vocabulary and embedding matrix, which are trained on the skill-centered dataset while keeping the model parameters frozen. 
}
\label{fig:fig2}
\vspace{-5pt}
\end{figure*}
\subsection{Skill-based Continual Learning: Problem Formulation}
\label{sec:problem_formulation}
Given the composition capabilities of modern LLMs~\citep{arora2023theory, yu2024skill, zheng2024enhancing} as well as their in-context learning abilities~\citep{wei2022emergent}, we investigate the following question: can LLMs learn new composable skills without weight updates?
We term this objective \textit{skill-based continual learning}, as it would allow models to acquire new composable skills without risk of catastrophic forgetting.
For such an approach to be practical and scalable, it requires
three key properties:

\begin{itemize}[leftmargin=3mm, itemsep=0pt, topsep=0pt]
    \item \textbf{Property 1 (No weight updates)}: New skills are learned without 
    modifying model parameters, preventing any catastrophic forgetting. 
    \item \textbf{Property 2 (Compositional transfer)}: Learned skills compose 
    with the model's existing skills in combinations not seen during training, 
    including out-of-distribution skill combinations.
    \item \textbf{Property 3 (Multi-skill composition)}: Multiple independently 
    learned skills can be composed together zero-shot, without joint training on their combination.
\end{itemize}

Property 2 is necessary for the learned skill to be composable with skills held by the model beyond the training distribution, while Property 3 enables scalable continual learning where skills 
can be added incrementally and composed together even without joint training.

We propose \textit{skill neologisms}—soft tokens integrated in the model's 
vocabulary—as one path towards achieving these properties. 
Our key hypothesis is that vocabulary-level interventions combined with skill-centered datasets can leverage the model's existing compositional abilities to learn composable representations of specific skills from the model's context.

\subsection{Skill Neologisms: Overview}
Skill neologisms are \textit{soft tokens} integrated in the model vocabulary and optimized such that providing them in the model's context enhances the model's capability for a specific skill.

Figure~\ref{fig:fig2} provides an overview of the different components required. We assume that a pretrained model $\mM$ has mastered a set of skills $\mS$ and has the ability to compose them (Figure~\ref{fig:fig2}A). Our aim is to learn a new skill $S^*$. First, a skill-centered dataset $\mD$ is constructed for skill $S^*$, with samples that all require at least skill $S^*$, as well as other skills from $\mS$ (Figure~\ref{fig:fig2}B). A skill neologism is initialized and added to the model's vocabulary and embeddings matrix. Then for each sample in $\mD$, the neologism is inserted in the prompt and trained on $\mD$ while keeping the rest of the model parameters fixed (Figure~\ref{fig:fig2}C).

\subsection{Skill-centered datasets}
\label{sec:skill-centered_dataset}
Most datasets used for model pretraining or fine-tuning are \textit{task-centered}: different samples or snippets of text implicitly depends on various skills. In contrast, skill neologisms require training on a dataset where every sample requires \textit{at least} the target skill, mixed with other skills mastered by the model (Figure~\ref{fig:fig1}).

\begin{definition}[Skill-centered dataset]
For a skill $S$ and set of skills $\mS$, an $S$-centered dataset is 
$\mD(S, \mS) = \{t_i \sim \mC_{k_i}(S, \mS)\}_i$, 
where each text snippet $t_i$ requires skill $S$ plus $k_i-1$ additional skills 
sampled from $\mS$, with $k_i$ drawn from $\{1, \ldots, k_{\max}\}$ 
according to some distribution $p$. We extend this notation to 
$(S_1, \ldots, S_m)$-centered datasets, where each snippet requires all of 
$S_1, \ldots, S_m$ plus up to $k_{\max}-m$ additional skills from $\mS$.
\end{definition}

\textbf{How to construct skill-centered datasets?}
Since the skills that underlie samples are usually implicit, it may not be immediately obvious how to construct datasets centered around a specific skill. However, we note that this is possible in many practical settings. First, in structured or synthetic settings, the mapping between samples and skills is often explicit by construction. For example in the experiments presented Section~\ref{sec:digit_exp}, each sample (e.g. \texttt{[ASC][ADD]4165=2567}) maps naturally to the underlying skills (e.g. \texttt{[ASC]} and \texttt{[ADD]}).
For more general settings, one can leverage the metacognitive abilities of strong LLMs to annotate samples with the implicit skills required~\citep{didolkar2024metacognitive}, and then filter to examples that require at least the skill of interest. Finally, many datasets provide expertly curated multi-labels categorizing each data entry--such as in educational problem banks~\citep{wang2020instructions, liu2023xes3g5m}, programming benchmarks~\citep{li2023taco}, or reasoning tasks~\citep{yuan2025mme}-- which can be used to filter data around specific skills.

\subsection{Skill Neologisms}

\begin{definition}[Skill neologism]
\label{def:neologism}
Given a model $\mM$ with parameters $\theta_\text{LLM}$, a \textit{skill neologism} for skill $S$ is a set of learnable soft tokens (or \textit{skill tokens}) $\theta_S \in \mathbb{R}^{d_{\text{model}} \times l}$ that minimizes some loss $\mathcal{L}$ over an $S$-centered dataset $\mD(S,\mS)$:
$$\theta_S^*=\argmin_{\theta_S} \mathbb{E}_{t \sim \mD(S, \Sigma)}[\mathcal{L}(\mathcal{M}(\theta_{\text{LLM}}, \theta_S, \phi_S(t)))]$$
where $\phi_S: \text{Text} \to \text{Text}$ is an \textit{insertion function} that inserts the skill neologism tokens into the text in a semantically appropriate way.
\end{definition}

The loss function $\mathcal{L}$ depends on the training paradigm: cross-entropy loss for supervised fine-tuning (SFT), RL-style objectives (e.g., policy gradients) for reinforcement fine-tuning (RFT), or preference-based losses as in \citet{hewittposition}. In our experiments, we use cross-entropy loss.

The choice of insertion function $\phi_S$ depends on the nature of the skill and how it naturally appears in text. We illustrate this with several examples below.

\textbf{Example of insertion functions}
Depending on the underlying skill and text snippet, the insertion function might be simply replacing a given word with the neologism $s$ --as done in~\citep{hewittposition} by replacing \textit{Ensure} by $\textit{Ensure}^h_w$--, or introducing a short text such as "Make sure to use $\langle$S$\rangle$" (see Table~\ref{tab:insertion} for examples under different settings).

\begin{table}[h]
\centering
\caption{Examples of neologism insertion functions $\phi_s$. In each case the tokens corresponding to the skill neologism are shown in $\langle$.$\rangle$ brackets.
}
\label{tab:insertion}
\resizebox{\linewidth}{!}{
\begin{tabular}{lll}
\toprule
\textbf{Setting} & \textbf{Original text $t$} & \textbf{Modified text $\phi_S(t)$} \\
\midrule
\makecell[l]{Word replacement \\ \citep{hewittposition}} & \makecell[l]{"Ensure that the length of the \\ response is at least 600 words."} & \makecell[l]{"$\langle$$\text{Ensure}^{h}_w$$\rangle$ that the length of the \\ response is at least 600 words."} \\
\makecell[l]{Word replacement \\ (Section~\ref{sec:digit_exp})} & "\texttt{[ADD][SHIFT]7283=...}" & "\texttt{[ADD]$\langle$SHIFT$\rangle$7283=...}" \\
Task instruction & "Sort these numbers:" & "Sort these numbers using $\langle$SORT$\rangle$:" \\
\bottomrule
\end{tabular}
}
\end{table}

\textbf{Training procedure.} 
We outline the training procedure for skill neologisms in Algorithm 1. 
\begin{algorithm}[h]
\scriptsize
\caption{Training Skill Neologisms}
\label{alg:skill_neologism}
\begin{algorithmic}[1]
\REQUIRE Pretrained model $\mM$ with frozen parameters $\theta_{\text{LLM}}$
\REQUIRE Target skill $S$, set of pretrained skills $\Sigma$, skill-centered dataset $\mathcal{D}(S, \Sigma)$
\REQUIRE Skill neologism length $l$, insertion function $\phi_{S}$
\REQUIRE Learning rate $\eta$, number of epochs $T$

\STATE Initialize skill tokens $\theta_{S} \in \mathbb{R}^{d_{\text{model}} \times l}$ 
\STATE Extend vocabulary: $\mathcal{V}' \leftarrow \mathcal{V} \cup \{\langle S_1 \rangle, \ldots, \langle S_l \rangle\}$
\STATE Extend embedding matrix: $E' \leftarrow E \cup \theta_{S}$

\FOR{$epoch = 1$ to $T$}
    \FOR{each batch $\mathcal{B} \subset \mathcal{D}(S, \Sigma)$}
        \FOR{each text sample $t \in \mathcal{B}$}
            \STATE $t' \leftarrow \phi_{S}(t)$ \hfill \# Insert skill tokens $\langle S_i \rangle$ into text
            \STATE Compute loss: $\mathcal{L} \leftarrow \text{CrossEntropy}(\mM(\theta_{\text{LLM}}, \theta_{S}, t'))$
            \STATE Compute gradients: $\nabla_{\theta_{S}} \mathcal{L}$
        \ENDFOR
        \STATE Update skill tokens: $\theta_{S} \leftarrow \theta_{S} - \eta \nabla_{\theta_{S}} \mathcal{L}$
        \STATE Keep model parameters $\theta_{\text{LLM}}$ unchanged
    \ENDFOR
\ENDFOR
\STATE \textbf{return} Optimized skill neologism $\theta_{S}^*$, extended vocabulary $\mathcal{V}'$
\end{algorithmic}
\end{algorithm}

\begin{figure*}[!t]
    \centering
    \includegraphics[width=1\linewidth]{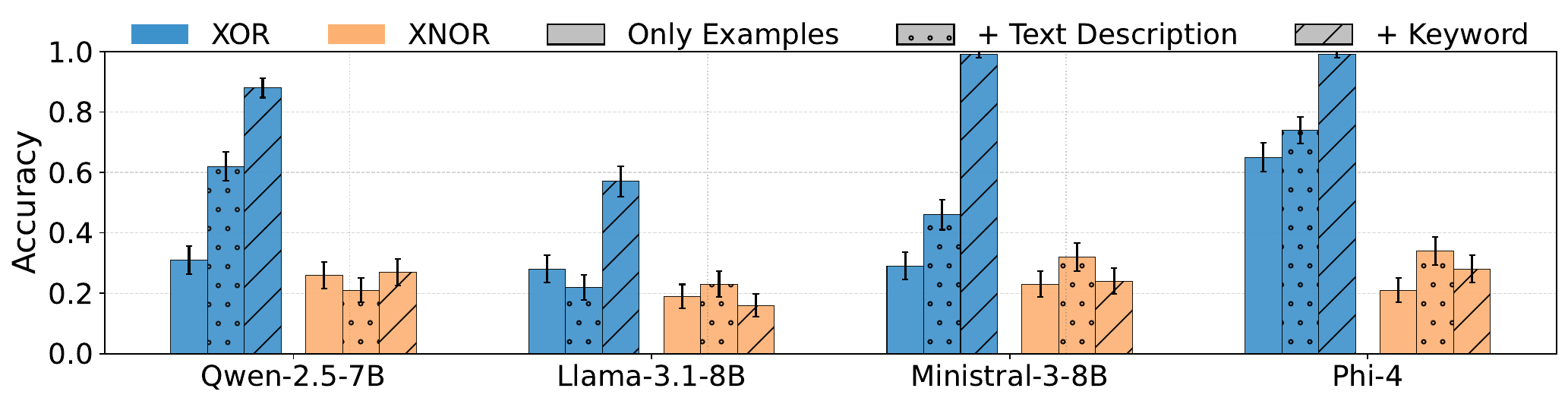}
    \vspace{-20pt}
    \caption{\textbf{Accuracy on XOR and XNOR completion tasks across open-source models under different prompts.} \textit{Only Examples} provides three input–output examples before the query. \textit{+ Text Description} adds a natural-language description of the operation (e.g., "output 1 iff the input bits differ" for XOR). \textit{+ Keyword} adds only the operation name ("XOR" or "XNOR"). Results are averaged over $N=100$ samples; error bars show standard error.}
    \label{fig:xor-xnor-accuracy}
    \vspace{-10pt}
\end{figure*}

\textbf{Evaluating compositional transfer.} 
After training a skill neologism, we assess whether it has learned a general representation of skill $S$ rather than only fitting the compositions between $S$ and skills from $\Sigma_\text{train}$. We adopt the notion of competence from~\citet{arora2023theory} where a model's competence $\tau_S$ on a skill $S$ is its success rate on an $S$-centered dataset. We evaluate the neologism on two datasets: (i) in-distribution (ID) combinations involving skills $\Sigma_\text{train}$ denoted as $\tau^{\text{ID}}_S$; (ii) out-of-distribution (OOD) combinations involving held-out skills $\Sigma_\text{test}$, denoted as $\tau^{\text{OOD}}_S$.

A successful neologism should achieve $\tau^{\text{OOD}}_S \approx \tau^{\text{ID}}_S$, indicating that the skill neologisms compose with novel skills zero-shot.

\subsection{Comparison to Existing Approaches}
Common approaches to extend model capabilities, such as LoRA and prompt tuning-like methods, are typically trained on task-centric datasets. As a result, they learn task-specific patterns rather than generally composable skills, limiting out-of-distribution transferability (P2). Moreover, these approaches are structurally unable to achieve P3: independently trained adapters or prefixes cannot be combined without retraining on the target task. 

Skill neologisms address this through two key components. First, skill-centered training with limited parameter capacity creates an inductive bias for learning generally composable skill representations (P2). Second, vocabulary-level integration leverages the model's in-context compositional abilities, allowing multiple independently learned skills to be combined simply by inserting multiple skill tokens in the context (P3).

%% file: sections/04-motivating_exp.tex
\section{Existence of Skill Tokens in Pretrained LLMs}
\label{sec:xor}
Before training skill neologisms in Section~\ref{sec:digit_exp}, we first illustrate that pretrained LLMs already exhibit analogous behavior—some vocabulary tokens are associated with specific procedural knowledge. During pretraining, certain tokens are encountered in contexts related to particular operations. For example, "\texttt{XOR}" tokens will frequently appear in text discussing the corresponding logical operation, which is analogous to a skill-centered dataset on the skill \texttt{XOR}. Consequently, these tokens might capture procedural knowledge for this operation. In contrast, less common tokens like "\texttt{XNOR}" might not—according to Google NGram Viewer, "\texttt{XOR}" appears approximately 15 times more frequently than "\texttt{XNOR}" in text from the past decade. We test this hypothesis on various open-source LLMs below.

\textbf{Setup} To test this hypothesis, we evaluate various open-source models on binary operation tasks. Models must perform \texttt{XOR} or \texttt{XNOR} on 3-bit sequences with 3 in-context examples. We compare accuracy across three conditions: (1) \textit{Only examples}; (2) \textit{Examples + description}: a textual description of the operation (e.g., ``\texttt{output 1 if and only if both input bits are different, and 0 otherwise}'' for \texttt{XOR}); and (3) \textit{Examples + keyword}: only the keyword "\texttt{XOR}" or "\texttt{XNOR}". For conditions (2) and (3), information is inserted before examples via: ``\texttt{Complete the following using the skill:{description/keyword}}''. Each model is evaluated on $N=100$ samples per setting.

\textbf{Results} Results are shown in Figure~\ref{fig:xor-xnor-accuracy}. For \texttt{XOR}, providing the keyword substantially improves accuracy over both other conditions, suggesting the "\texttt{XOR}" token has captured procedural knowledge through pretraining exposure, functioning as a genuine skill token. In contrast, for \texttt{XNOR}, neither keyword nor description improves accuracy beyond examples alone, indicating the ``\texttt{XNOR}'' token lacks sufficient training signal to encapsulate this operation. This demonstrates that skill tokens can emerge naturally when vocabulary tokens have sufficient exposure to skill-relevant contexts, and that they can encode procedural knowledge more efficiently than explicit descriptions.

\vspace{-5pt}
\begin{highlight}{\textbf{Takeaway}}
Pretrained LLMs exhibit tokens that encode procedural knowledge, motivating the use of vocabulary-level parameters to learn skill representations.
\end{highlight}
\vspace{-5pt}

%% file: sections/05_experiments.tex
\section{Algorithmic Skill Composition}
In this section, we evaluate skill neologisms on a controlled digit-sequence transformation task. We chose this setting because it provides explicit sample-skill definitions and unambiguous composition rules, unlike natural language tasks where skills are typically implicit. This enables us to construct exact ID/OOD splits over skills to cleanly measure whether the model learns general representations that compose with held-out skills (P2) and whether independently trained skills can be combined zero-shot (P3).

\label{sec:digit_exp}
\subsection{Setup}
\textbf{Dataset} We create a synthetic dataset based on operations over digits sequences. Each sample is of the form: "$\texttt{[OP-1]}\ldots\texttt{[OP-k]x=y}$", where \texttt{x} is a random sequence of n digits, each \texttt{OP-i} is an operation and the output is the result of sequentially applying operations to x: $\texttt{y}=(\texttt{OP-k} \circ \ldots \circ \texttt{OP-1})(\texttt{x})$. Table~\ref{tab:ops} shows the different operations and example samples for $n=3$. 

\begin{table}[h]
\centering
\caption{Digit-sequence transformation skills.}
\label{tab:ops}
\resizebox{\linewidth}{!}{
\begin{tabular}{c l l l}
\hline
\textbf{Set} & \textbf{Skill} & \textbf{Description} & \textbf{Example (Seq. length: 3)} \\
\hline
\multirow{7}{*}{\Large\(\mSpt\)}
 & \texttt{ASC} & Sort digits in ascending order & \([\texttt{ASC}]472 = 247\) \\
 & \texttt{DESC} & Sort digits in descending order & \([\texttt{DESC}]472 = 742\) \\
 & \texttt{ADD} & Add 1 to each digit & \([\texttt{ADD}]472 = 583\) \\
 & \texttt{SUB} & Subtract 1 from each digit & \([\texttt{SUB}]472 = 361\) \\
 & \texttt{REV} & Reverse digit order & \([\texttt{REV}]472 = 274\) \\
 & \texttt{POL} & Map odd (even) digits to 1 (0) & \([\texttt{POL}]472 = 010\) \\
 & \texttt{ID} & Identity mapping & \([\texttt{ID}]472 = 472\) \\
\hline
\multirow{2}{*}{\Large\(\Snew\)}
 & \texttt{SHIFT} & Right-shift digits & \([\texttt{SHIFT}]472 = 247\) \\
 & \texttt{INV-POL} & Map odd (even) digits to 0 (1) & \([\texttt{INV-POL}]472 = 101\) \\
\hline
\end{tabular}
}
\end{table}

\textbf{Base model} We fine-tune Qwen2.5-0.5B~\citep{qwen2025qwen25technicalreport} on $\mD(\mSpt)$ with up to 3-skill combinations, using digit sequences of lengths $n \in [2,9] \setminus \{5,7,9\}$, holding out lengths 5, 7, and 9 for validation. To ensure the model learns to combine operations flexibly, we also hold out 25\% of 3-skill combinations. Training uses LoRA~\citep{hu2022lora} in two phases: (i) 100k single-skill samples and (ii) 500k samples with $k=\{1,2,3\}$ drawn uniformly. Table~\ref{tab:pt_3ops_run15} shows accuracy across skill counts and sequence lengths. The model achieves high accuracy on both in-distribution and held-out lengths, and generalizes well to unseen 3-skill combinations—with the exception of \texttt{REV}, which we exclude from $\mS_{\mathrm{held}\text{-}\mathrm{out}}$ due to overfitting on held-out lengths (see Appendix~\ref{app:pretrain_details} and~\ref{appendix:pretraining}).

\begin{table}[t]
\centering
\caption{Accuracy of $\mathcal{M}_{pretrain}$ over sequence lengths for single-task (\(\mathcal{C}_1\)), two-task (\(\mathcal{C}_2\)), and three-task (\(\mathcal{C}_3\)) compositions. * indicates lengths and combinations held-out during training. ID: in-distribution skill combinations, OOD: out-of-distribution skill combinations. Per-operation details shown in Appendix~\ref{app:pretrain_details}.}
\resizebox{\linewidth}{!}{\input{tables/pretrain_3ops}}
\end{table}

\textbf{Learning new skills} 
We then freeze the pretrained model $\mM_\mathrm{pretrain}$ and aim to learn two new skills $\Sigma_\mathrm{test}=\{\texttt{SHIFT}, \texttt{INV-POL}\}$. For each skill $S_\mathrm{new}$, we generate a dataset of 100k samples with 1-, 2-, and 3-combinations of $S_\mathrm{new} \cup \Sigma_\mathrm{train}$. To test out-of-distribution generalization, we create multiple leave-one-out datasets by setting $\Sigma_{train}=\Sigma_\mathrm{pretrain} \setminus \Sho$.
This allows to test in-distribution on $\Sigma_{train}$ and out-of-distribution on $\Sho$.

\textbf{Models} For each new skill $\Snew \in \Sigma_{\text{test}}$ and held-out skill $\Sho \in \Sigma_{\text{pretrain}}$, we train three model variants on 100k samples from $D(\Snew, \Sigma_{\text{pretrain}} \setminus \Sho)$ with up to $k_\mathrm{max}=3$ operations: (1) \textbf{Skill neologisms} with length $l=20$, initialized from the mean embedding of $\Sigma_{\text{pretrain}}$ operation tokens; (2) \textbf{Prompt tuning}~\citep{lester2021power} with a prefix of length $l=20$ using the same initialization; (3) \textbf{LoRA}~\citep{hu2022lora} with rank $r=16$.

\subsection{Results}
\label{sec:result_main}

\begin{figure*}[t]
    \centering
    \includegraphics[width=1\linewidth]{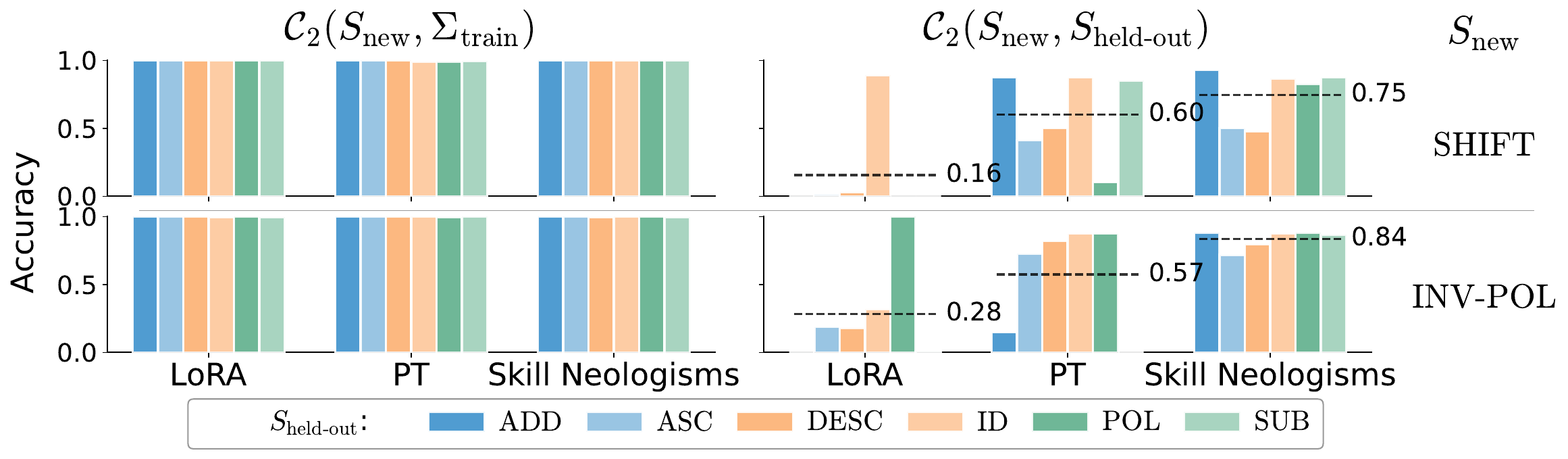}
    \vspace{-15pt}
    \caption{Accuracy on 2-combinations of skills mixing $\Snew$ with $\mStrain$ (in-distribution) or $\Sho$ (out-of-distribution). Dotted lines show the average accuracy across all $\Sho$. PT: Prompt Tuning.}
    \label{fig:zero_shot_exp}
\end{figure*}

We now validate that skill neologisms satisfy P2 and P3 from Section~\ref{sec:problem_formulation}. P1 (no weight updates) is satisfied by construction, as all model parameters are frozen when training skill neologisms. We evaluate P2 by testing whether learned skills compose with held-out skills not seen during training, and P3 by testing whether independently learned skill neologisms can be combined zero-shot.

\subsubsection{Property 2: Compositional Transfer}
\label{sec:result_p2}
Figure~\ref{fig:zero_shot_exp} shows the accuracy of LoRA, prompt tuning, and skill neologisms on 2-combinations of $\Snew$ with either skills from $\mStrain$ (in-distribution) or $\Sho$ (out-of-distribution). All three methods achieve near-perfect in-distribution accuracy. However, only skill neologisms consistently succeed at composing $\Snew$ with $\Sho$. LoRA shows the poorest OOD generalization, suggesting it overfits the training distribution rather than learning a composable representation of $\Snew$. Prompt tuning performs intermediately; the gap with skill neologisms is notable given both optimize the same number of soft tokens. This suggests that semantically embedding the soft tokens inside the prompts may provide additional flexibility to learn composable skill representation. Accuracy on 3-combinations shows similar patterns (Figure~\ref{app:fig:zero_shot_exp_k3} in Appendix).

\begin{highlight}{\textbf{Takeaway}}
Skill neologisms learn composable skill representations that transfer to OOD compositions.
\end{highlight}

\subsubsection{Property 3: Multi-Skill Composition}
\label{sec:multi-skill}
We test whether the skill neologisms learned independently for \texttt{SHIFT} and \texttt{INV-POL} in \S~\ref{sec:result_p2} can be combined zero-shot to handle compositions requiring both skills (Property 3). This distinguishes skill neologisms from LoRA and prompt tuning-like approaches, which cannot be composed after independent training without retraining on the joint task. We compare against in-context learning (ICL)--a natural baseline for zero-shot composition--by providing $\mMpt$ with $N \in \{10, 20, 50, 100\}$ examples from $\mD(S_\mathrm{new}, \mSpt)$ for $S_\mathrm{new}\in\{\texttt{SHIFT},\texttt{INV-POL}\}$ ($2N$ examples in total).

\begin{figure}[h]
    \centering
    \includegraphics[width=1\linewidth]{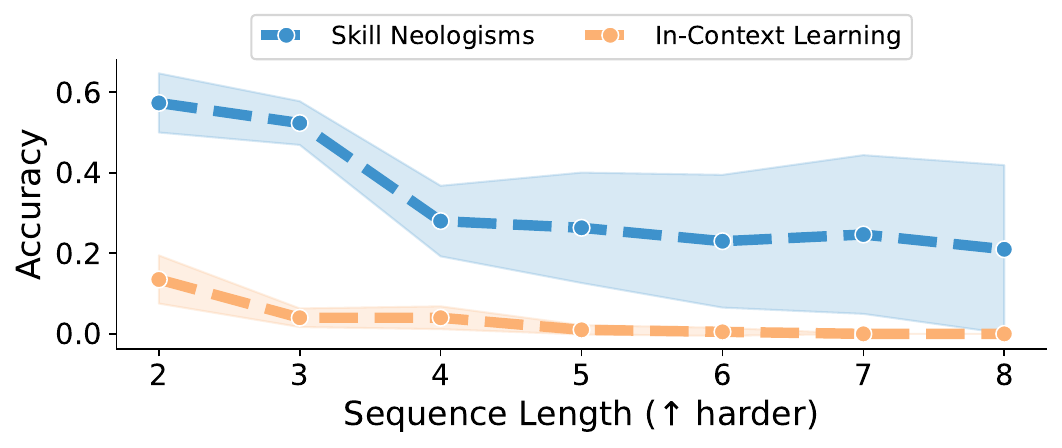}
    \caption{\textbf{Zero-shot composition of \texttt{SHIFT} and \texttt{INV-POL}}. \textit{Skill Neologisms}: we compose the skill tokens learned independently for \texttt{SHIFT} and \texttt{INV-POL} for a given $\Sho$, and plot the average accuracy (±std) across the 6 $\Sho$. 
    \textit{In-Context Learning}: we provide in-context $N=\{10, 20, 50, 100\}$ examples sampled from $\mD(S_\mathrm{new}, \mSpt)$ for $S_\mathrm{new}\in\{\texttt{SHIFT},\texttt{INV-POL}\}$ ($2N$ examples in total), and plot the average accuracy(±std) across the 4 runs.
    }
    \label{fig:p3_results}
\end{figure}

Figure~\ref{fig:p3_results} shows the average accuracy across different $\Sho$ for skill neologisms and $N$ for ICL, for different sequence lengths (increasing task difficulty). Traces for individual runs are shown Figure~\ref{app:fig:result_p3_detailed} in the Appendix. Skill neologisms significantly outperform ICL across all sequence lengths. This demonstrates that skill neologisms successfully capture reusable procedural knowledge that transfers zero-shot to new compositions, whereas ICL struggles to extract and combine the relevant patterns from examples alone.

\begin{highlight}{\textbf{Takeaway}}
Independently learned skill neologisms can be successfully composed zero-shot.
\end{highlight}

\subsection{Insights and Ablation Experiments}
\label{sec:result_ablation}
Having validated that skill neologisms satisfy Properties 2 and 3, we study and ablate different components to understand the mechanisms at play. We focus on three questions: (1) How does the capacity of skill tokens affect their ability to learn composable representations? (2) How does the diversity of skill combinations in training data impact generalization? (3) Is performance sensitive to initialization and label noise?

\subsubsection{Skill Neologism Length}
\textbf{Motivation} We hypothesize that limited capacity of skill tokens provides an inductive bias to learn generally composable representations rather than overfitting to the training distribution, as we observed with LoRA in Figure~\ref{fig:zero_shot_exp}. This suggests a trade-off on the length of the skill neologism: too few parameters may fail to learn the skill, while too many may reduce generalization to OOD compositions.

\textbf{Setup} We train skill neologisms for \texttt{SHIFT} and \texttt{INV-POL} with $\Sho=\texttt{ADD}$, varying the neologism length from $l=1$ ($|\theta_S|=768$ parameters) to $l=200$ ($|\theta_S|=153$k parameters).

\begin{figure}[h]
    \centering
    \includegraphics[width=1\linewidth]{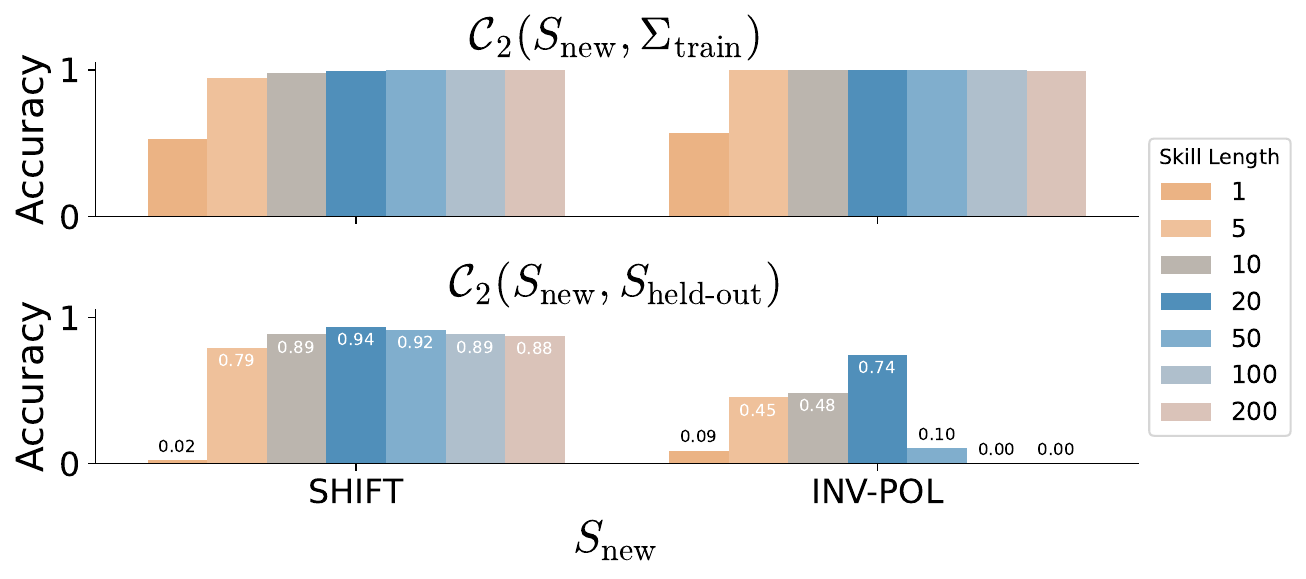}
    \caption{\textbf{Effect of skill token length.} Accuracy on 2-skill combinations with $\Sho=\texttt{ADD}$ for varying skill token length $l$.}
    \label{fig:effect_length}
\end{figure}

\textbf{Results} Figure~\ref{fig:effect_length} shows the accuracy on 2-combinations with $\mStrain$ (ID) and $\Sho$ (OOD) for varying skill lengths. The model gets near-perfect accuracy in-distribution for $l\geq5$. However, the accuracy out-of-distribution first increases with higher capacity, but then drops as $l$ becomes too large ($l>20$). This suggests that after a certain point, increased capacity for the skill tokens becomes detrimental to learning a generally composable representation of the skill.

\begin{highlight}{\textbf{Takeaway}}
The limited capacity of skill neologisms acts as an inductive bias to learn more composable skill representations. 
\end{highlight}

\subsubsection{Composition Complexity in Training Set}
\label{sec:compo_complex}
The complexity of skill combinations in the training set provides another source of inductive bias: exposing skill tokens to the target skill in more complex compositions (higher $k_{\max}$) may improve their ability to compose with held-out skills.

\textbf{Setup} We train neologisms for \texttt{SHIFT} and \texttt{INV-POL} for various $\Sho$, while varying the maximum number of compositions $k_{\max} \in \{1,2,3\}$ in the training set, keeping the total number of samples fixed at 100k. We compare the OOD accuracy on 2- and 3-compositions involving the held-out skill.

\textbf{Results} Table~\ref{tab:vark} shows the accuracy averaged across all $\Sho$ (see Appendix~\ref{app:sec:vark_detailed} for detailed results across $\Sho$). Training on more compositions in the training data generally improves OOD generalization. In particular, 2-skill compositions benefit from having been trained on 3-skill composition data for \texttt{INV-POL}.

\begin{table}[h]
\centering
\caption{\textbf{Effect of composition complexity in the training set.} OOD accuracy(±std) on 2-skill and 3-skill combinations when training with varying maximum composition complexity $k_{\max}$. Results averaged across all held-out skills $\Sho$ (see Appendix~\ref{app:sec:vark_detailed} for a detailed breakdown across $\Sho$).}
\label{tab:vark}
\resizebox{\linewidth}{!}{\input{tables/effect_of_kop_training_table-l20}}
\end{table}

\subsubsection{Robustness to Initialization and Label Noise}
\label{sec:init}
\textbf{Effect of initialization} Prompt tuning-like methods are known to depend on initialization~\citep{lester2021power}. Table~\ref{tab:init_method} compares random initialization against initialization from the average embedding of tokens in $\mSpt$ (see Appendix~\ref{app:sec:init_detailed} for detailed results across $\Sho$). Initialization from pretrained skill embeddings shows marginally better performance (particularly for \texttt{INV-POL} on 2-skill compositions), but skill tokens trained from random initialization still show strong OOD composition abilities.

\begin{table}[h]
\centering
\caption{\textbf{Effect of initialization.} OOD accuracy(±std) on 2-skill and 3-skill combinations for random initialization versus initialization from average embeddings of pretrained skills $\mSpt$. Results averaged across all held-out skills $\Sho$.}
\label{tab:init_method}
\resizebox{\linewidth}{!}{\input{tables/effect_of_init_method_table-l20-avg}}
\end{table}

\textbf{Effect of label noise} 
Accurately curating skill-centered datasets might not always be possible, resulting in noisy skill labels. 
To evaluate the impact of noise on skill labels, we train skill neologisms for \texttt{INV-POL} with $\Sho=\texttt{ADD}$ while randomly replacing occurrences of the target skill in the training set with a random other skill and updating the ground-truth answer accordingly. This results in the skill neologism being trained on a dataset that sometimes involves an operation that is not the expected skill. The results are presented Figure~\ref{fig:skill_label_noise} in the Appendix, and show that while the 1-op accuracy is robust up to around $40\%$ of erroneous labels, noise in the skill labels degrades 2-op and 3-op performance with $20\%$ noise, particularly for OOD compositions.

\begin{figure*}[t!]
\centering
\includegraphics[width=1\linewidth]{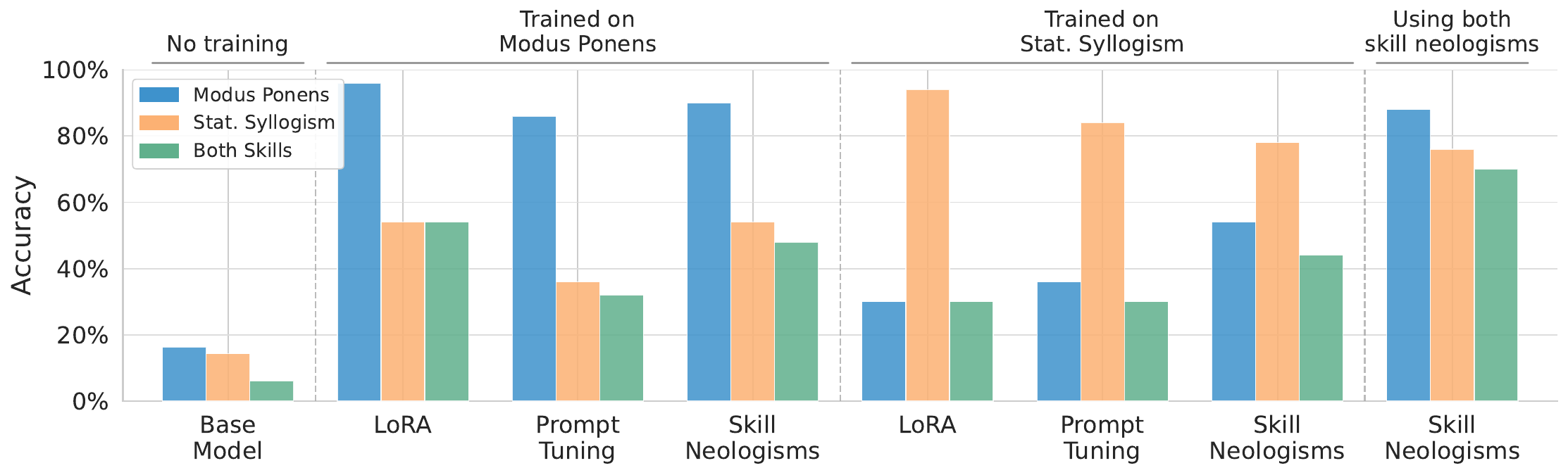}
\caption{\textbf{Learning two target skills on Skill-Mix}. “No training”: vanilla Llama3.2-3B-Instruct model. “Trained on $S_{new}$”: each model is trained on a $S_{new}$-centered dataset mixing $S_{new}$ with other skills. “Using both skill neologisms”: combining the independently learned skill tokens zero-shot. "Both Skills" corresponds to successfully producing a text that includes both skills. Grading is done with GPT-5. Prompts for inference and grading follow the official Skill-Mix repository.}
\vspace{-5pt}
\label{fig:skillmix}
\end{figure*}

\section{Natural Language Skill Composition}
\label{sec:skill-mix}
To demonstrate the applicability of skill neologisms to a realistic language setting, we conduct experiments on the Skill-Mix benchmark~\citep{yu2024skill}, which evaluates the ability of LLMs to combine natural language skills. For each sample, the model is asked to generate a short text that illustrates a list of language skills (e.g. "metaphor" or "syllogism") on a specific topic. Then, the produced text is graded by a strong LLM (here, GPT-5) based on whether or not it includes the desired skills. While this benchmark was introduced with the aim to measure LLMs' ability to combine skills, it provides a useful testbed to evaluate a model's mastery and acquisition of different skills. In our case, we use this setting to demonstrate that neologisms learned independently for different skills can be successfully composed zero-shot.

\textbf{Setup} We run Llama3.2-3B-Instruct on the Skill-Mix repository (containing 10 publicly released skills) and select two skills for which the base model has low accuracy: “modus ponens” and “statistical syllogism”. For each target skill, we construct N=300 training samples by mixing the target skill with 0, 1 or 2 randomly sampled skills (while holding out the other target skill) and collecting high-quality answers using GPT-5. We train with LoRA, prompt tuning and skill neologisms via SFT and test on 50 queries requiring both target skills simultaneously. We use the prompts from the Skill-Mix repository, which include a definition and one example for each queried skill (detailed hyperparameters and prompts are provided in Appendix~\ref{app:sec:skillmix}). For skill neologisms, we use $l=20$ and insert neologisms into the prompt by replacing every occurrence of the skill name with the corresponding soft tokens.

\textbf{Results} Figure~\ref{fig:skillmix} shows the accuracy of each method after being trained on either one of the target skills. LoRA, prompt tuning and skill neologisms reach similarly high accuracy for the skill they were trained for, but not for the other held-out target skill. Skill neologisms allow for the zero-shot combination of both independently learned skill tokens, achieving high accuracy on both skills simultaneously despite never being trained on them jointly.

%% file: tables/pretrain_3ops.tex
\label{tab:pt_3ops_run15}
\begin{tabular}{c c c c c}
\toprule
 & \multicolumn{4}{c}{\textbf{Composition setting}} \\
\cmidrule(lr){2-5}
\textbf{Sequence length} 
& \(\mathcal{C}_1(\mSpt)\) 
& \(\mathcal{C}_2(\mSpt)\) 
& \multicolumn{2}{c}{\(\mathcal{C}_3(\mSpt)\)} \\
\cmidrule(lr){4-5}
 & \textbf{ID} & \textbf{ID} & \textbf{ID} & \textbf{OOD*} \\
\midrule
2 & 100.0\% & 100.0\% & 100.0\% & 97.0\% \\
3 & 100.0\% & 100.0\% & 100.0\% & 96.0\% \\
4 & 99.1\%  & 100.0\% & 98.0\%  & 97.0\% \\
5* & 97.6\%  & 98.0\%  & 95.0\%  & 84.0\% \\
6 & 95.6\%  & 94.0\%  & 95.0\%  & 89.0\% \\
7* & 92.6\%  & 92.0\%  & 89.0\%  & 74.0\% \\
8 & 92.6\%  & 90.0\%  & 79.0\%  & 74.0\% \\
9* & 83.9\%  & 75.0\%  & 74.0\%  & 58.0\% \\
\bottomrule
\end{tabular}

%% file: tables/effect_of_kop_training_table-l20.tex
\begin{tabular}{llcc}
\toprule
$S_{new}$ & $k_\mathrm{max}$ & $\mathcal{C}_2(S_{new}, \Sho)$ & $\mathcal{C}_3(S_{new}, \Sho, \mStrain)$ \\
\midrule
\multirow[t]{3}{*}{\texttt{INV-POL}} 
&1 & .36 $\pm$ .44 & .05 $\pm$ .06 \\
&2 & .60 $\pm$ .34 & .61 $\pm$ .24 \\
&3 & .85 $\pm$ .06 & .65 $\pm$ .23 \\
\cline{1-4}
\multirow[t]{3}{*}{\texttt{SHIFT}} 
&1 & .46 $\pm$ .04 & .55 $\pm$ .04 \\
&2 & .72 $\pm$ .21 & .61 $\pm$ .19 \\
&3 & .72 $\pm$ .18 & .69 $\pm$ .14 \\
\bottomrule
\end{tabular}

%% file: tables/effect_of_init_method_table-l20-avg.tex
\begin{tabular}{llcc}
\toprule
$S_{new}$ & Init Method & $\mathcal{C}_2(S_{new}, \Sho)$ & $\mathcal{C}_3(S_{new}, \Sho, \mStrain)$ \\
\midrule
$\texttt{INV-POL}$ & From $\mSpt$ & .85 $\pm$ .06 & .65 $\pm$ .23 \\
$\texttt{INV-POL}$ & Random & .63 $\pm$ .32 & .58 $\pm$ .31 \\
\midrule
$\texttt{SHIFT}$ & From $\mSpt$ & .72 $\pm$ .18 & .69 $\pm$ .14 \\
$\texttt{SHIFT}$ & Random & .70 $\pm$ .19 & .65 $\pm$ .14 \\
\bottomrule
\end{tabular}

%% file: sections/06_discussion.tex
\vspace{-5pt}
\section{Discussion}
\label{sec:discussion}
\textbf{Related Work} Our work relates to three main research directions (detailed comparison in Appendix~\ref{app:sec:related_work}). First, prior research has investigated skills and compositional abilities in LLMs using in-context skill descriptions~\citep{chen2023skills}, skill-rich synthetic data~\citep{zhao2024can, kaur2025instruct}, or skill-targeted training~\citep{he2025stat}. In contrast, we learn generally composable skill representations via soft tokens integrated into the model vocabulary. Second, while prefix-based adaptation methods~\citep{lester2021power, li2021prefix} and their transferability extensions~\citep{vu2022spot, wang2023multitask} improve generalization across tasks, they ultimately require training on the target task. We instead adopt a skill-centric perspective, focusing on out-of-distribution generalization and zero-shot composition of independently learned skills. Finally, meaningful soft tokens have been studied for visual concepts~\citep{gal2023image}, tool representations~\citep{hao2023toolkengpt}, prompt compression~\citep{mu2023learning,sastre2025memory,kuratov2025cramming}, and behavioral alignment~\citep{radevski2026compositional}. To the best of our knowledge, our work is the first to propose learning composable soft tokens that encapsulate specific procedural knowledge.

\textbf{Limitations and Future Work} Our work constitutes an initial proof-of-concept of skill neologisms as a path towards skill-based continual learning, focusing on a controlled experimental setting and a natural language task where skill can be well-defined. Further work is needed to explore skill neologisms in other settings where the definition of skills and their compositions is fuzzier. Key challenges include the construction and availability of diverse skill-centered datasets, as well as the optimization instability inherent to training soft tokens (see detailed discussion in Appendix~\ref{app:sec:lim}). 

\textbf{Conclusion} We propose skill neologisms as a way to extend LLM capabilities to specific skills by optimizing vocabulary-integrated soft tokens on skill-centered data. By demonstrating compositional transfer to out-of-distribution skills and zero-shot combination of independently learned soft tokens, both in a controlled algorithmic task and a more realistic natural language setting, our findings confirm that skill neologisms are a promising direction for scalable skill-based continual learning.

%% file: sections/app_00_main.tex
\input{sections/app_01_additional_experiments}

\input{sections/app_02_related_work}

\input{sections/app_03_experiment_details}

\makeatletter 
\renewcommand{\thefigure}{\@arabic\c@figure}
\makeatother

%% file: sections/app_01_additional_experiments.tex
\section{Extended Results}
\subsection{Model pre-training}

\begin{figure}[h]
    \centering
    \includegraphics[width=1\linewidth]{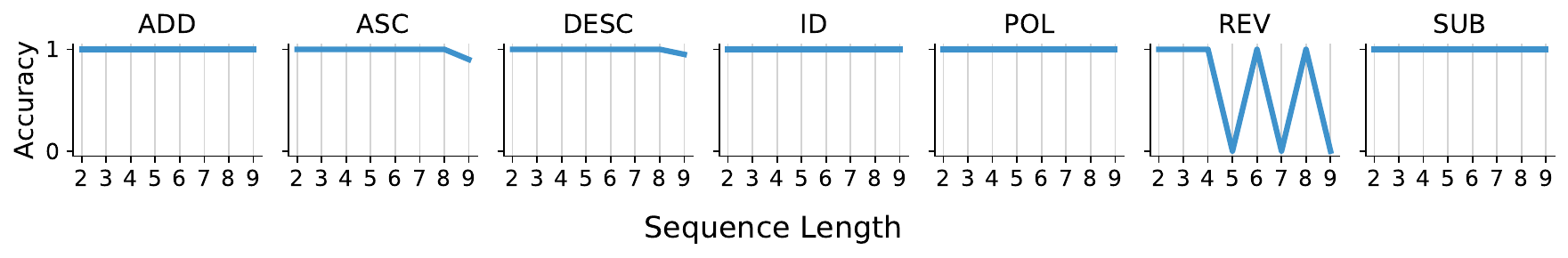}
    \vspace{-15pt}
    \caption{Accuracy of $\mMpt$ across sequence lengths for each pre-train operation. Sequence lengths $\{2,3,4,6,8\}$ are in-distribution, while lengths $\{5, 7, 9\}$ were held-out from pre-training data.}
    \label{app:pretrain_details}
\end{figure}

Figure~\ref{app:pretrain_details} shows the accuracy of $\mMpt$ after pre-training (same as Table~\ref{tab:pt_3ops_run15}), across sequence lengths and operations. Sequence lengths $\{2,3,4,6,8\}$ are in-distribution, while lengths $\{5, 7, 9\}$ were held-out from pre-training data. The model successfully learns most operations over the training distribution and generalizes to unseen sequence lengths. The only exception is \texttt{REV}, which does not generalize to OOD sequence lengths.

\subsection{Out-of-distribution generalization}
\begin{figure*}[h!]
    \centering
    \includegraphics[width=1\linewidth]{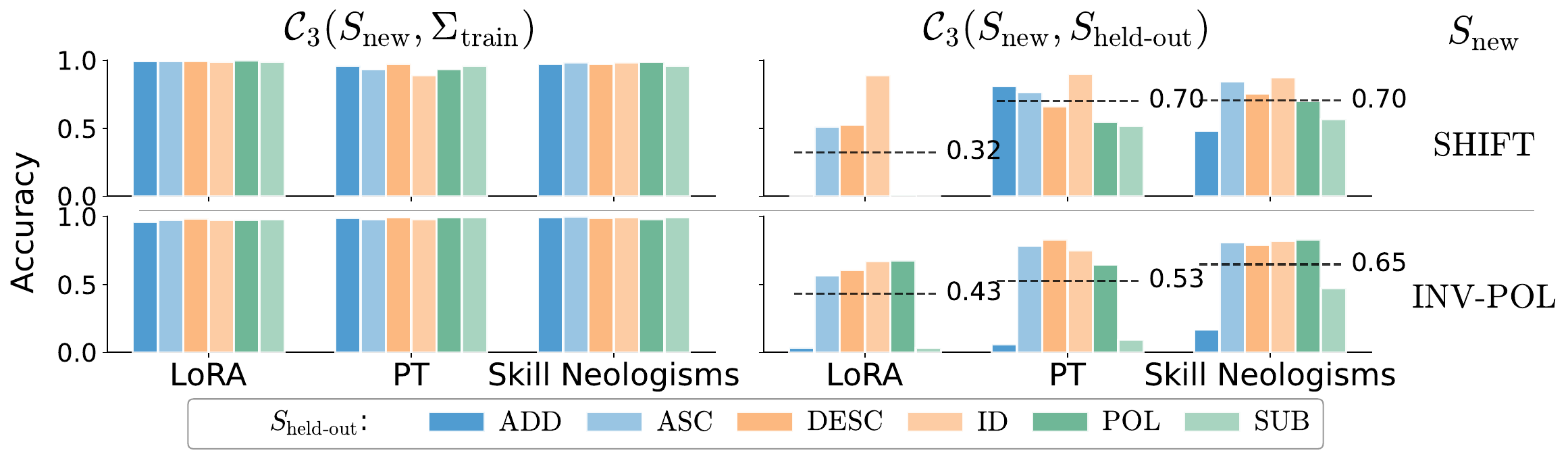}
    \vspace{-15pt}
    \caption{Accuracy on 3-combinations of skills mixing $\Snew$ with $\mStrain$ (in-distribution) or $\Sho$ (out-of-distribution). Dotted lines show the average accuracy across all $\Sho$. PT: Prompt Tuning.}
    \label{app:fig:zero_shot_exp_k3}
\end{figure*}

Following the experimental setup from Section~\ref{sec:result_p2}, Figure~\ref{app:fig:zero_shot_exp_k3} shows the ID and OOD accuracy on 3-compositions of skills. For OOD, samples are drawn from $\mathcal{C}_3(S_{new}, \Sho, \mStrain)$, where $\Snew$ and $\Sho$ are always included and one operation from $\mStrain$ is sampled, and the order of the three operations is randomly permuted. 

\subsection{Multi-skill composition}
\begin{figure*}[h!]
    \centering
    \includegraphics[width=0.7\linewidth]{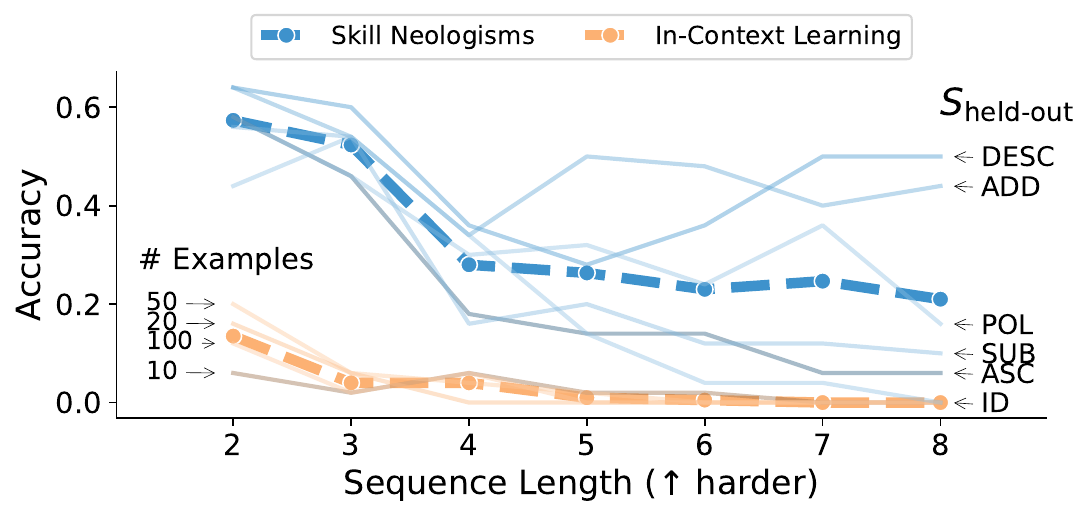}
    \caption{\textbf{Zero-shot composition of \texttt{SHIFT} and \texttt{INV-POL}}. \textit{Skill Neologisms}: we compose the skill tokens learned independently for \texttt{SHIFT} and \texttt{INV-POL} for a given $\Sho$ (thin blue lines), and plot the average accuracy (±std) across the 6 $\Sho$ (thick dashed blue line). 
    \textit{In-Context Learning}: we provide in-context $N=\{10, 20, 50, 100\}$ examples sampled from $\mD(S_\mathrm{new}, \mSpt)$ for $S_\mathrm{new}\in\{\texttt{SHIFT},\texttt{INV-POL}\}$ ($2N$ examples in total), and plot individual results (thin orange lines) and the average (thick dashed orange line) across the 4 runs.
    Thin lines show the individual runs across $\Sho$ and $N$.}
    \label{app:fig:result_p3_detailed}
\end{figure*}

Figure~\ref{app:fig:result_p3_detailed} shows detailed results from Section~\ref{sec:multi-skill} with accuracy on individual pairs of neologisms (for a given $\Sho$) for skill neologisms, and individual number of examples $N$ for ICL.

\subsection{Effect of compositions in training set}
\label{app:sec:vark_detailed}
Table~\ref{app:tab:vark_detailed} shows the detailed accuracy across $\Sho$ skills for the experiment presented Section~\ref{sec:compo_complex}.

\begin{table*}[h!]
\centering
\caption{\textbf{Effect of number of compositions in training set.}}
\label{app:tab:vark_detailed}
\resizebox{\linewidth}{!}{\input{tables/effect_of_kop_training_table-l20-detailed}}
\end{table*}

\subsection{Effect of initialization}
\label{app:sec:init_detailed}
Table~\ref{app:tab:init_detailed} shows the detailed accuracy across $\Sho$ skills for the initialization ablation presented Section~\ref{sec:init}.

\begin{table}[h!]
\vspace{-10pt}
\centering
\caption{\textbf{Effect of initialization.} OOD accuracy on 2-skill and 3-skill combinations for random initialization versus initialization from average embeddings of pretrained skills $\mSpt$.}
\label{app:tab:init_detailed}
\resizebox{\linewidth}{!}{\input{tables/effect_of_init_method_table-l20-detailed}}
\end{table}

\subsection{Effect of noisy skill labels}
\label{app:sec:skill_label_noise}
\begin{figure}[]
\centering
\includegraphics[width=0.8\linewidth]{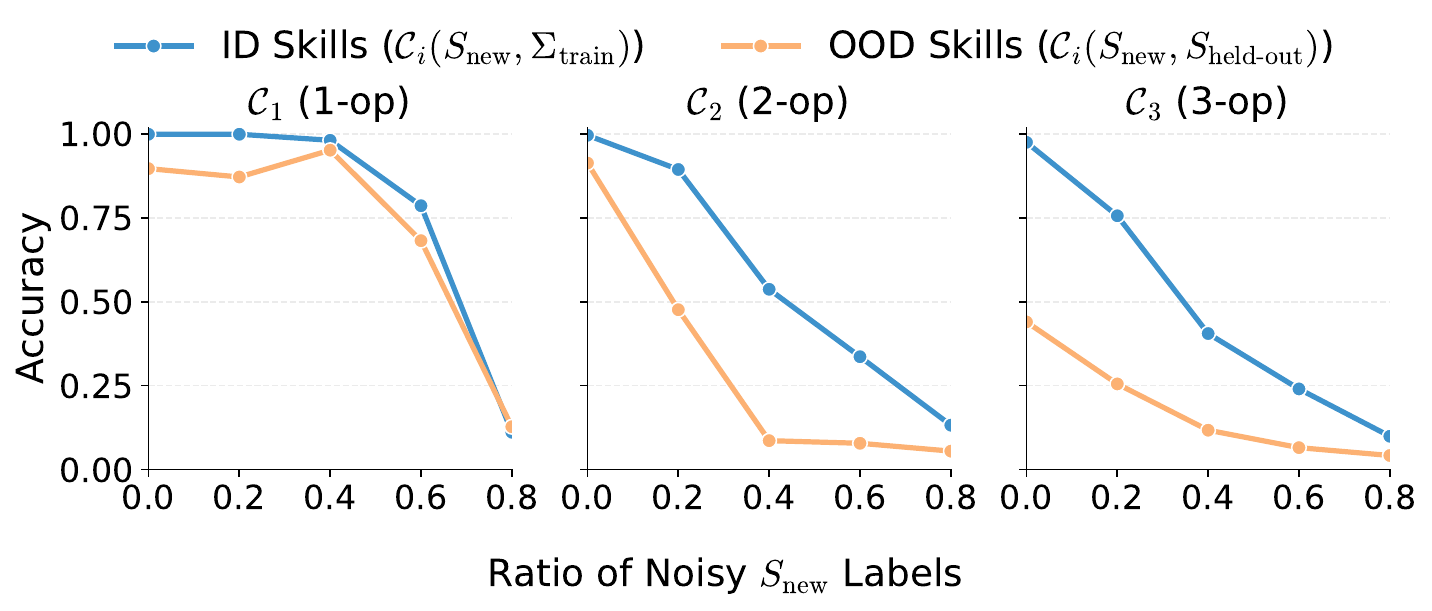}
\caption{\textbf{Effect of noise in the skill-centered train set.} Underlying operations labeled as the target skills $S_{new}$ are corrupted to another operation with varying proportions in the training set. Using parameters $l=5$, $S_{new}=\texttt{INV-POL}$ and $S_{\mathrm{held-out}}=\texttt{ADD}$.
Increasing label noise on the target skill gradually degrades performance, particularly for 2- and 3-skill compositions and OOD skills.
}
\label{fig:skill_label_noise}
\end{figure}

Figure~\ref{fig:skill_label_noise} shows the impact of varying magnitudes of noise on skill labels on downstream accuracy, using $l=5$, $S_{new}=\texttt{INV-POL}$ and $S_{\mathrm{held-out}}=\texttt{ADD}$.

%% file: tables/effect_of_kop_training_table-l20-detailed.tex
\begin{tabular}{llcccccccccccccc}
\toprule
 &  & \multicolumn{7}{c}{Acc on $\mathcal{C}_2(S_{new}, \Sho)$} & \multicolumn{7}{c}{Acc on $\mathcal{C}_3(S_{new}, \Sho, \mSpt)$} \\ \cmidrule(lr){3-9} \cmidrule(lr){10-16}
 &  & ASC & DESC & ADD & SUB & ID & POL & \textbf{AVG} & ASC & DESC & ADD & SUB & ID & POL & \textbf{AVG} \\
$S_{new}$ & k-ops &  &  &  &  &  &  &  &  &  &  &  &  &  &  \\
\midrule
\multirow[t]{3}{*}{INV-POL} & 1 & .15 & .06 & .00 & .00 & .99 & .95 & .36 & .07 & .03 & .00 & .00 & .02 & .18 & .05 \\
 & 2 & .43 & .80 & .46 & .00 & .90 & .98 & .60 & .76 & .81 & .33 & .22 & .83 & .73 & .61 \\
 & 3 & .73 & .80 & .88 & .86 & .89 & .91 & .85 & .83 & .78 & .21 & .47 & .81 & .82 & .65 \\
\cline{1-16}
\multirow[t]{3}{*}{SHIFT} & 1 & .48 & .51 & .40 & .42 & .46 & .51 & .46 & .58 & .62 & .51 & .53 & .53 & .56 & .55 \\
 & 2 & .40 & .50 & .95 & .92 & .84 & .71 & .72 & .69 & .76 & .38 & .37 & .90 & .58 & .61 \\
 & 3 & .52 & .46 & .92 & .87 & .87 & .70 & .72 & .82 & .73 & .49 & .54 & .88 & .70 & .69 \\
\cline{1-16}
\bottomrule
\end{tabular}

%% file: tables/effect_of_init_method_table-l20-detailed.tex
\begin{tabular}{llcccccccccccccc}
\toprule
 &  & \multicolumn{7}{r}{Acc on $\mathcal{C}_2(S_{new}, \Sho)$} & \multicolumn{7}{r}{Acc on $\mathcal{C}_3(S_{new}, \Sho, \mSpt)$} \\
 &  & \texttt{ADD} & \texttt{ASC} & \texttt{DESC} & \texttt{ID} & \texttt{POL} & \texttt{SUB} & \textbf{AVG} & \texttt{ADD} & \texttt{ASC} & \texttt{DESC} & \texttt{ID} & \texttt{POL} & \texttt{SUB} & \textbf{AVG} \\
$S_{new}$ & Init Method &  &  &  &  &  &  &  &  &  &  &  &  &  &  \\
\midrule
\multirow[t]{2}{*}{\texttt{INV-POL}} & From $\mathcal{S}_{pretrain}$ & .88 & .73 & .80 & .89 & .91 & .86 & .85 & .21 & .83 & .78 & .81 & .82 & .47 & .65 \\
 & Random & .44 & .80 & .81 & .88 & .88 & .00 & .63 & .23 & .80 & .79 & .78 & .83 & .05 & .58 \\
\cline{1-16}
\multirow[t]{2}{*}{\texttt{SHIFT}} & From $\mathcal{S}_{pretrain}$ & .92 & .52 & .46 & .87 & .70 & .87 & .72 & .49 & .82 & .73 & .88 & .70 & .54 & .69 \\
 & Random & .88 & .46 & .41 & .87 & .70 & .85 & .70 & .50 & .77 & .71 & .87 & .54 & .50 & .65 \\
\cline{1-16}
\bottomrule
\end{tabular}

%% file: sections/app_02_related_work.tex
\section{Extended Related Work}
\label{app:sec:related_work}
\textbf{Skills and compositional abilities in LLMs}
Recent works have proposed ways to extend model capabilities to new skills and compositions. 
Skill-in-Context~\citep{chen2023skills} aims to elicit compositional abilities in LLMs by providing in-context descriptions of skills and step-by-step explanations on how to compose them. ~\citet{zhao2024can} show that training LLMs on skill-rich synthetic datasets improve compositional abilities, even on held-out skills unseen during training. STAT~\citep{he2025stat} aims to improve model capabilities by uncovering specific skills lacking from the model, and targeting these skills via either reweighting or synthetic data augmentations. \citet{didolkar2024metacognitive} demonstrated that LLMs have the ability to describe skills required by a given task, while~\citet{kaur2025instruct} leveraged such metacognition abilities of LLMs to create a skill-rich synthetic dataset for instruction-tuning. In contrast, we propose learning generally composable representation of skills via soft tokens, allowing composition with other skills thanks to the pretrained model's general compositional abilities.

\textbf{Prefix-Based Adaptations}
Prompt tuning~\citep{lester2021power} first introduced the paradigm of training soft tokens appended to the input prompt to adapt a frozen model to new tasks, which P-Tuning~\citep{liu2024ptuning} extended by mixing soft prompts produced by a prompt encoder with discrete text tokens. Concurrently, prefix tuning~\citep{li2021prefix} proposed learning prefix key and value vectors at every layer of the model--yielding more expressive power than the input layer only--which P-Tuning v2~\citep{liu2022p} extended to natural language understanding (NLU) settings. Several works have focused on enhancing the transferability of prompt tuning. SPoT~\citep{vu2022spot} train prompts across diverse tasks to improve transferability to new ones; Multitask Prompt Tuning (MTP)~\citep{wang2023multitask} decomposes prompts between shared and task-specific components; ATTEMPT~\citep{asai2022attempt} combines prompts from different tasks using an attention mechanism. However, these methods still require training on the target task, unlike skill neologisms that can combine independently learned soft-prompts zero-shot.

\textbf{Meaningful Soft Tokens}  
Another line of research has focused on learning soft tokens with specific, grounded meanings, moving beyond their use as purely task-specific adapters. In the vision-language domain, Textual Inversion~\citep{gal2023image} learns a new pseudo-word in the embedding space of a frozen model to represent a novel visual concept, such as a specific object or artistic style. In function calling and tool use for LLMs, ToolkenGPT~\citep{hao2023toolkengpt} represents tools via tokens integrated in the model vocabulary. In prompt compression, memory tokens~\citep{sastre2025memory, kuratov2025cramming} compress long sequences of text into a single reversible embedding, while gist tokens~\citep{mu2023learning} replace prompts with gist tokens that preserve downstream model behavior. 
Recently, \citet{radevski2026compositional} proposed learning composable steering tokens for behavioral alignment. 
To the best of our knowledge, our work is the first to learn composable soft tokens that encapsulate specific procedural knowledge.

\section{Extended Limitations} 
\label{app:sec:lim}
\paragraph{Skill-centered dataset construction}
While we argue that skill-centered datasets can be identified or constructed in a variety of contexts, their availability for a given skill remains a key requirement for learning skill neologisms. Moreover, as our experiments suggest, the quality of the learned neologisms partly depends on the complexity of the data and on how the target skill is mixed with a diverse set of other skills during training. Assessing and ensuring such diversity may not be straightforward in all settings.

\paragraph{Soft token limitations}
Skill neologisms rely on optimizing soft tokens, in a manner similar to prompt tuning. As a result, they inherit several limitations commonly associated with prompt tuning, including sensitivity to initialization and to hyperparameters such as token length and learning rate. In addition, successfully learning soft tokens requires that the target task remains reasonably close to the model’s pretraining distribution, as shown in~\citet{petrov2024prompting} and~\citet{genewein2025understanding}.

\paragraph{Computational cost}
Although soft tokens are substantially more parameter-efficient than standard fine-tuning methods, training them still requires backpropagation through the full model. This leads to computational costs that can be comparable to those of fine-tuning in practice. As a result, training skill neologisms for large-scale models (e.g., $>30$B parameters) may remain challenging without access to substantial computational resources.

%% file: sections/app_03_experiment_details.tex
\newpage
\section{Experimental Details}
Code to reproduce our experiments can be found at {\url{https://github.com/antoninbrthn/skill-neologisms}} and \url{https://github.com/vanderschaarlab/skill-neologisms}.

\subsection{Experimental Details: Section 4}
\label{appendix:xor-experiments}

\subsubsection{Experimental Setup}

\textbf{Models Evaluated:}
\begin{itemize}
    \item Qwen2.5-7B~\citep{qwen2025qwen25technicalreport}
    \item Llama-3.1-8B~\citep{grattafiori2024llama}
    \item Ministral-3-8B-Base-2512~\citep{liu2026ministral}
    \item Phi-4~\citep{abdin2024phi}
\end{itemize}

\textbf{Tasks:} Binary operations XOR and XNOR on 3-bit sequences.

\textbf{Dataset Configuration:}
\begin{itemize}
    \item Test samples: 100 per task (XOR, XNOR)
    \item Bit length: 3
    \item In-context examples: 3 examples per prompt
    \item Example format: Each sample contains 3 input-output pairs followed by a query input
\end{itemize}

\textbf{Prompt Variations:} Three prompt formulations were tested for each task:

\begin{enumerate}
    \item \textbf{Only Examples (Baseline):} No additional context provided, only the 3 in-context example pairs
    
    \item \textbf{Examples + Keyword:} A symbolic keyword prefix is added before the examples
    \begin{itemize}
        \item XOR: ``Complete the following using the skill: `XOR' ''
        \item XNOR: ``Complete the following using the skill: `XNOR' ''
    \end{itemize}
    
    \item \textbf{Examples + Text Description:} A natural language description is provided
    \begin{itemize}
        \item XOR: ``Complete the following using the skill: `output 1 iif both input bits are different, and 0 otherwise' ''
        \item XNOR: ``Complete the following using the skill: `output 1 iif both input bits are the same, and 0 otherwise' ''
    \end{itemize}
\end{enumerate}

\textbf{Example Prompt Structure:}

For the ``Examples + Keyword'' variant (XOR):
\begin{verbatim}
Complete the following using the skill: 'XOR'
101 011 = 110
100 110 = 010
011 001 = 010
111 010 = 
\end{verbatim}

For the ``Only Examples'' variant:
\begin{verbatim}
101 011 = 110
100 110 = 010
011 001 = 010
111 010 = 
\end{verbatim}

For the ``Examples + Text Description'' variant (XOR):
\begin{verbatim}
Complete the following using the skill: 'output 1 iif 
both input bits are different, and 0 otherwise'
101 011 = 110
100 110 = 010
011 001 = 010
111 010 = 
\end{verbatim}

\subsubsection{Evaluation Details}

\textbf{Inference Parameters:}
\begin{itemize}
    \item Batch size: 16
    \item Generation method: Greedy decoding (deterministic)
    \item Padding side: left
    \item Models run in evaluation mode
\end{itemize}

\textbf{Metrics:}
\begin{itemize}
    \item \textbf{Exact Match Accuracy:} Percentage of test samples where the model's generated output exactly matches the ground truth
    \item \textbf{Standard Error:} Computed assuming binomial distribution: $\text{SE} = \sqrt{\frac{p(1-p)}{N}}$ where $p$ is accuracy and $N=100$
\end{itemize}

\subsection{Experimental Details: Section 5}

This appendix provides comprehensive details for all experiments presented in Section~\ref{sec:digit_exp}.

\subsubsection{Base Model Pretraining}
\label{appendix:pretraining}

All experiments in Section~\ref{sec:digit_exp} use a pretrained Qwen2.5-0.5B model trained on a digit-sequence transformation task. Table~\ref{tab:pretraining} summarizes the pretraining configuration. To avoid biasing the model to expect operations in specific input positions, we prepend each training sample with a random number of padding tokens, within a limit of $128$ tokens per sample. 

\begin{table}[h]
\centering
\caption{Base model pretraining configuration. The model was trained in two phases: Phase 1 on single operations, Phase 2 on compositions of 1--3 operations.}
\label{tab:pretraining}
\small
\begin{tabular}{lcc}
\toprule
\textbf{Parameter} & \textbf{Phase 1} & \textbf{Phase 2} \\
\midrule
Base Model & \multicolumn{2}{c}{Qwen/Qwen2.5-0.5B} \\
PEFT Method & \multicolumn{2}{c}{LoRA (r=32, $\alpha$=32)} \\
Target Modules & \multicolumn{2}{c}{q, k, v, o, gate, up, down} \\
\midrule
Training Samples & 100,000 & 500,000 \\
Test Samples & 500 & 500 \\
Operations per Sample & 1 & 1--3 \\
Epochs & 3 & 3 \\
\midrule
Batch Size & 64 & 64 \\
Learning Rate & 2e-4 & 2e-4 \\
Warmup Steps & 500 & 500 \\
\midrule
Operations & \multicolumn{2}{c}{\texttt{[ASC]}, \texttt{[DESC]}, \texttt{[ADD]}, \texttt{[SUB]},} \\
              & \multicolumn{2}{c}{\texttt{[POL]}, \texttt{[REV]}, \texttt{[ID]}} \\
Sequence Lengths & \multicolumn{2}{c}{2, 3, 4, 6, 8 (held-out: 5, 7, 9)} \\
\midrule
Held-out 3-op combinations & -- & 25\% \\
\bottomrule
\end{tabular}
\end{table}

\subsubsection{Skill Neologisms}
\paragraph{Insertion function} In Section~\ref{sec:digit_exp}, the insertion function $\phi$ simply swaps the tokens corresponding to the target skill (e.g. "\texttt{[SHIFT]}") with the skill tokens of length $l$ in the prompt.

\subsubsection{Compositional Transfer Experiments (Figure 4)}
\label{appendix:compositional-transfer}

Table~\ref{tab:methods-comparison} summarizes the configuration for each method in Figure 4.

\begin{table}[h]
\centering
\caption{Configuration for compositional transfer experiments (Figure 4). All methods learn one of \texttt{[SHIFT]} and \texttt{[INV-POL]} and are evaluated on compositions with held-out pretrain operations.}
\label{tab:methods-comparison}
\small
\begin{tabular}{lccc}
\toprule
\textbf{Parameter} & \textbf{Skill Neologisms} & \textbf{Prompt Tuning} & \textbf{LoRA} \\
\midrule
Trainable Structure & Vocab. tokens & Prefix tokens & LoRA adapters \\
Soft Tokens Length/Rank & 20 & 20 & r=16 \\
Trainable Params & 17,920 & 17,920 & $\sim$2.9M \\
\midrule
Initialization & \multicolumn{2}{c}{Mean of pretrain op. embeddings} & -- \\
\midrule
Training Samples & \multicolumn{3}{c}{100,000} \\
Validation Samples & \multicolumn{3}{c}{1,000} \\
Test Samples & \multicolumn{3}{c}{200 per Sequence Length and permutation} \\
Operations per Sample & \multicolumn{3}{c}{1--3 (requires $S_{new}$ + 0--2 from $\Sigma_{train}$)} \\
Held-out Skill & \multicolumn{3}{c}{One operation per run (6 total scenarios)} \\
Sequence Lengths & \multicolumn{3}{c}{2, 3, 4, 6, 8 (held-out: 5, 7, 9)} \\
\midrule
Epochs & 3 & 3 & 3 \\
Learning Rate & 5e-3 & 5e-3 & 1e-4 \\
Batch Size & 32 & 32 & 32 \\
Temperature at Inference & \multicolumn{3}{c}{0 (greedy)} \\
\midrule
Eval. Metrics & \multicolumn{3}{c}{Acc. on $\mathcal{C}_2(S_{new}, \Sigma_{train})$ (ID)} \\
              & \multicolumn{3}{c}{Acc. on $\mathcal{C}_2(S_{new}, S_{held\text{-}out})$ (OOD)} \\
\bottomrule
\end{tabular}
\end{table}

\textbf{Dataset Configuration:} Training samples compose $S_{new}$ with operations from $\Sigma_{train}$ (6 of the 7 pretrain operations, with one held out). Training and validation data is distributed equally across operation counts (e.g., for max\_ops=3, each of 1-op, 2-op, and 3-op receives $\frac{100{,}000}{3} \approx 33{,}333$ samples).

\textbf{Test Dataset Generation:} The test dataset evaluates all permutations of operation orderings to ensure the model learns composable skills rather than memorizing specific sequences. For each $k \in \{2, 3\}$ operations:
\begin{itemize}
    \item Each sample requires exactly one $S_{new}$, one $S_{held\text{-}out}$, and $(k-2)$ operations from $\Sigma_{train}$
    \item The order of these 2 (resp. 3) operations is set by sampling one of the 2 (resp. 6) permutations of $S_{new}$, $S_{held\text{-}out}$, and $S\in \Sigma_{train}$.
    \item $N_{test}=200$ samples are generated for each sequence length and permutations, yielding 400 test samples per sequence length for k=2 and 1200 test samples per sequence length for k=3.
\end{itemize}

Each method is trained on 6 configurations (one per held-out operation) for each of the 2 new skills, yielding 12 runs per method.

\subsubsection{Multi-Skill Composition Experiments (Figure 5)}
\label{appendix:multiskill}

Table~\ref{tab:multiskill-config} presents the experimental setup for Figure 5. 

\begin{table}[h]
\centering
\caption{Configuration for multi-skill composition experiments (Figure 5). Skill neologisms for \texttt{[SHIFT]} and \texttt{[INV-POL]} are learned independently, then composed zero-shot.}
\label{tab:multiskill-config}
\small
\begin{tabular}{lp{9cm}}
\toprule
\textbf{Parameter} & \textbf{Value} \\
\midrule
\multicolumn{2}{l}{\textbf{Skill Neologisms}} \\
Training & Two independently trained skills (config from Table~\ref{tab:methods-comparison}) given a held-out skill $\Sho$, repeated for 6 different $\Sho$ \\
Composition Method & Insert both skill tokens into test prompts \\
Evaluation & Repeat across all 6 $\Sho$ \\
\midrule
\multicolumn{2}{l}{\textbf{In-Context Learning Baseline}} \\
Examples per Skill & $N \in \{10, 20, 50, 100\}$ \\
Total Examples & $2N$ (N for each target skill) \\
Examples Pool Size & 10,000 samples per skill \\
\midrule
\multicolumn{2}{l}{\textbf{Test Dataset}} \\
Test Samples & 50 per sequence length \\
Sequence Lengths & 2--8 \\
Operations per Sample & \texttt{[SHIFT]} and \texttt{[INV-POL]} \\
Temperature & 1 \\
\bottomrule
\end{tabular}
\end{table}

\subsubsection{Ablation Studies}
\label{appendix:ablations}

\paragraph{Effect of Training Composition Complexity}

Table~\ref{tab:kops-ablation} shows how varying the maximum number of operations during training (max\_ops) affects generalization. Other parameters are the same as in Section~\ref{appendix:compositional-transfer} under the "Skill Neologisms" column. 

\begin{table}[h]
\centering
\caption{Effect of training composition complexity. Each row shows results for a different max\_ops value during training. All configurations use skill length 20.}
\label{tab:kops-ablation}
\small
\begin{tabular}{lcccc}
\toprule
\textbf{max\_ops} & \textbf{Epochs} & \textbf{Training Samples} \\
\midrule
1,2,3 & 2 & 100,000 \\
\bottomrule
\end{tabular}
\end{table}

\textbf{Evaluation:} Each configuration is evaluated on both 2-operation and 3-operation compositions with held-out skills. The table in the paper reports mean accuracy across all 6 held-out operations for each $S_{new}$.

\paragraph{Effect of Initialization Method}

Table~\ref{tab:init-ablation} compares initialization strategies for skill token embeddings. Other parameters are the same as in Section~\ref{appendix:compositional-transfer} under the "Skill Neologisms" column. 

\begin{table}[h]
\centering
\caption{Initialization method comparison. Both methods use skill length 20, learning rate 5e-3, and 2 epochs of training.}
\label{tab:init-ablation}
\small
\begin{tabular}{lcc}
\toprule
\textbf{Method} & \textbf{Description} \\
\midrule
From Pretrain & Mean of pretrain operation embeddings \\
Random  & Random Gaussian initialization with $\sigma=0.2$ \\
\bottomrule
\end{tabular}
\end{table}

\textbf{Evaluation:} Average accuracy on $\mathcal{C}_2$ and $\mathcal{C}_3$ compositions across all 6 held-out operations.

\paragraph{Effect of Skill Token Length (Figure 6)}

Figure 6 shows how skill token capacity affects learning and generalization. Parameters are the same as in Section~\ref{appendix:compositional-transfer} under the "Skill Neologisms" column, while only varying the skill token length $l\in\{1,5,10,20,50,100,200\}$.

\begin{table}[h]
\centering
\caption{Configuration for length ablation experiments.}
\label{tab:length-ablation-config}
\small
\begin{tabular}{lp{7cm}}
\toprule
\textbf{Parameter} & \textbf{Value} \\
\midrule
Skills Evaluated & \texttt{[SHIFT]}, \texttt{[INV-POL]} \\
Fixed Held-out Skill & \texttt{[ADD]} \\
Training Samples & 100,000 \\
max\_ops & 2 (1 or 2 operations per sample) \\
\midrule
Epochs & 1 \\
Learning Rate & 5e-3 \\
Batch Size & 32 \\
\midrule
Metrics & ID: Acc. on $\mathcal{C}_2(S_{new}, \Sigma_{train})$ \\
        & OOD: Acc. on $\mathcal{C}_2(S_{new}, \texttt{ADD})$ \\
\bottomrule
\end{tabular}
\end{table}




\subsubsection{Dataset and Evaluation Details}
\label{appendix:dataset-details}

\textbf{Operations:} All experiments use 7 pretrained operations on digit sequences:
\begin{itemize}
    \item \texttt{[ASC]}: Sort digits in ascending order
    \item \texttt{[DESC]}: Sort digits in descending order
    \item \texttt{[ADD]}: Add 1 to each digit (mod 10)
    \item \texttt{[SUB]}: Subtract 1 from each digit (mod 10)
    \item \texttt{[POLARITY]}: Map odd digits to 1, even to 0
    \item \texttt{[REVERSE]}: Reverse digit order
    \item \texttt{[ID]}: Identity (no transformation)
\end{itemize}

Two new operations are learned in all main experiments:
\begin{itemize}
    \item \texttt{[SHIFT]}: Right-shift digits cyclically
    \item \texttt{[INV-POL]}: Map odd digits to 0, even to 1
\end{itemize}

\textbf{Sequence Lengths:} 
\begin{itemize}
    \item Training: 2, 3, 4, 6, 8
    \item Held-out: 5, 7, 9
\end{itemize}

\textbf{Sample Format:} Each sample follows the pattern \texttt{[OP-1]...[OP-k]xxxx=yyyy}, where \texttt{xxxx} is the input digit sequence and \texttt{yyyy} is the result of applying operations sequentially.

\textbf{Evaluation Metrics:}
\begin{itemize}
    \item \textbf{Exact Match Accuracy:} The model must generate the complete correct output sequence.
\end{itemize}

\subsubsection{Computational Resources}
\label{appendix:compute}

\textbf{Model:} Qwen/Qwen2.5-0.5B
\begin{itemize}
    \item Embedding Dimension: 896
    \item Hidden Size: 896
    \item Layers: 24
    \item Attention Heads (Q / KV): 14 / 2
    \item Tie Embeddings: Yes
\end{itemize}

\textbf{Framework:}
\begin{itemize}
    \item HuggingFace Transformers
    \item PEFT library for LoRA
    \item Custom implementation for skill neologisms and prompt tuning (same implementation for both, simply inserting soft tokens before every prompt for prompt tuning)
    \item Weights \& Biases for experiment tracking
\end{itemize}

\textbf{Hardware:} Experiments were run on a NVIDIA RTX 6000 GPU (48GB VRAM).

\newpage

\subsection{Experimental Details: Section 6}
\label{app:sec:skillmix}
We run the Skill-mix benchmark~\citep{yu2024skill} based on the reference implementation\footnote{\url{https://github.com/LeoYu/skill-mix/}}.

\subsubsection{Skill-Mix Experiment: Modus Ponens and Statistical Syllogism}

We evaluate \texttt{meta-llama/Llama-3.2-3B-Instruct} on the public Skill-Mix benchmark using the target skills \texttt{modus ponens} and \texttt{statistical syllogism}. 

\paragraph{Training set construction.} The training data is constructed as follows. For $k=2,3$, we mix each target skill with a subset of skills from the publicly released skills in the Skill-mix repository. 

\begin{itemize}
    \item \textbf{Target skills:} \texttt{modus ponens} and \texttt{statistical syllogism}.
    \item \textbf{Training set size:} 300 GPT-5 generated samples per skill, split evenly across \(k=1,2,3\) with 100 samples each.
    \item \textbf{Train skills:} \texttt{self serving bias}, \texttt{red herring}, \texttt{spatial reasoning}, and \texttt{folk physics (common knowledge physics)}.
\end{itemize}

\paragraph{Method hyperparameters.} The method-specific training settings are:

\begin{itemize}
    \item \textbf{Skill Neologisms:} \(l=20\) soft tokens per skill; SFT for 10 epochs; learning rate \(5\times10^{-3}\); batch size 8; gradient accumulation 2; linear learning rate scheduler; random initialization.
    \item \textbf{Prompt Tuning:} same \(l=20\) tokens prefix length and hyperparameters as skill neologisms.
    \item \textbf{LoRA:} rank 16, LoRA alpha 16, dropout 0.1, targeting \texttt{q\_proj}, \texttt{k\_proj}, \texttt{v\_proj}, \texttt{o\_proj}, \texttt{gate\_proj}, \texttt{up\_proj}, and \texttt{down\_proj}.
\end{itemize}

\paragraph{Testing setup.} We evaluate on 50 test $k=2$ queries with both target skills jointly, using \texttt{temperature=1.0}, and GPT-5 with \texttt{temperature=1.0} and \texttt{reasoning\_effort=minimal} for the grading.

\subsubsection{Prompts}
We use the same prompts template for inference and grading as the original Skill-mix implementation, reproduced in Listings~\ref{lst:base_prompt} and~\ref{lst:grading}.

\begin{lstlisting}[caption={Prompt template for Skill-Mix queries.}, label={lst:base_prompt}]
Greetings! I am interested in natural language processing and I was wondering if you could help me generate an example of text that illustrates multiple skills in semantics or syntax. The example should be a minimal natural piece of text with up to a few lines in the context of {topic} that illustrates all of the following skills: {skills_str}. Please keep the text as short as possible, and make sure the concepts can be found fully from the text. 

For reference, here are the definitions and examples for the concepts:
{skills_definitions_and_examples}

Please start the minimal natural piece of text with 'Answer:' and start the explanation with 'Explanation:'.

Thanks very much!
\end{lstlisting}

\begin{lstlisting}[caption={Prompt template for Skill-Mix grading.}, label={lst:grading}]
Greetings! I was wondering if you could help me grade the following answer given by a student.

I'll first describe the question that was given to the student, and then give you the student's answer, and the grading rubric.

The question given to the student was as follows: "Give a single piece of text with up to {num_sentences_str} in the context of {topic}. This single piece of text should illustrate all of the following skills: {skills_str}."

The student's answer was: "{student_answer}"

For reference, here are the definitions for the skills:
{skills_defs_and_examples_simple}

Using a rubric table format, please grade the student's answer with positive scoring. Each criterion is worth 1 point. The criteria are: {rubric_items}. The table should only have the following columns: 'Criteria', 'Points Earned'. In the 'Points Earned' column, please provide only numerical values with no additional formatting. Please introduce the table with 'Here's the grading table:' and please include a row for 'Total Points Earned' at the end of the table. Finally, please start your grading explanation with 'Explanation':
\end{lstlisting}

Below is an example prompt with skills "modus ponens" and "statistical syllogism" on the topic of "gardening":

\begin{lstlisting}
Greetings! I am interested in natural language processing and I was wondering if you could help me generate an example of text that illustrates multiple skills in semantics or syntax. The example should be a minimal natural piece of text with up to a few lines in the context of Gardening that illustrates all of the following skills: modus ponens, statistical syllogism. Please keep the text as short as possible, and make sure the concepts can be found fully from the text. 

For reference, here are the definitions and examples for the concepts:
**modus ponens**: A syllogism that is of the form "If P then Q. P. Hence Q." For example, "If today is Tuesday, then John will go to work. Today is Tuesday. Therefore, John will go to work."
**statistical syllogism**: A syllogism that argues, using inductive reasoning, from a generalization true for the most part to a particular case. For example, "Almost all people are taller than 26 inches. Gareth is a person. Therefore, Gareth is taller than 26 inches."

Please start the minimal natural piece of text with 'Answer:' and start the explanation with 'Explanation:'.

Thanks very much!
\end{lstlisting}

For skill neologisms, we insert skill tokens at every location where the exact skill name appears in the prompt. For example, inserting neologisms of length $l=20$ for "modus ponens" and "statistical syllogism" in the prompt above results in the following input prompt, where $\texttt{<|skill\_id-i|>}$ denotes the i-th soft token for that skill neologism.

\begin{lstlisting}
Greetings! [...] the following skills: <|modus_ponens-0|>...<|modus_ponens-19|>, <|statistical_syllogism-0|>...<|statistical_syllogism-19|>. Please keep the text as short as possible, and make sure the concepts can be found fully from the text. 

For reference, here are the definitions and examples for the concepts:
**<|modus_ponens-0|>...<|modus_ponens-19|>**: [...]
**<|statistical_syllogism-0|>...<|statistical_syllogism-19|>**: [...]

Please start [...]
\end{lstlisting}